\documentclass[fleqn,10pt]{wlscirep}
\usepackage[utf8]{inputenc}
\usepackage[T1]{fontenc}

\usepackage{comment}
\usepackage{graphicx} 
\usepackage{todonotes}
\usepackage{amsmath,amssymb,amsfonts}
\usepackage{bm}

\usepackage{booktabs}
\usepackage{multirow}
\usepackage{amsmath, amssymb}
\usepackage{algorithmic}
\usepackage[linesnumbered,ruled,vlined]{algorithm2e}
\usepackage{color}
\usepackage[normalem]{ulem}
\usepackage{float}

\title{Identifiable Decomposition of Submovements in Human Hand Trajectories}

\author[1,*]{Adrian Prados}
\author[2,*]{James Hermus}
\author[1]{Ramon Barber}
\author[2,3]{Sylvain Calinon}
\affil[1]{Universidad Carlos III de Madrid, Leganes, Spain} 
\affil[2]{Idiap Research Institute, Martigny, Switzerland} 
\affil[3]{Ecole Polytechnique F{\'e}d{\'e}rale de Lausanne (EPFL), Lausanne, Switzerland} 

\affil[*]{aprados@ing.uc3m.es, james.hermus@idiap.ch}


\begin{abstract}
Voluntary movements have long been hypothesised to be comprised of discrete primitives called submovements, as a descriptive model of human motor behaviour. However, existing methods scale poorly, and no principled method exists to determine whether a decomposition is informative. We propose a spatiotemporal kernel correlation between primitive pairs as an identifiability criterion. Submovement-Identifiable Decomposition (Sub-ID) embeds this criterion in its adaptive-ridge regularisation, biasing the optimiser toward low-correlation solutions. Identifiability is lost when primitives become collinear and recovered when they diverge spatially. On synthetic data, Sub-ID recovers ground-truth parameter distributions where existing methods fail; furthermore, when primitives overlap too heavily to be distinguished, the method explicitly detects this ambiguity rather than outputting misleading results. Sub-ID extracts submovements from real three-dimensional, long-horizon movements --- a regime no prior method addresses. This method has the potential to identify physiologically grounded primitives for motor control research and imitation learning.
\end{abstract}
\begin{document}

\flushbottom
\maketitle
%
%
\thispagestyle{empty}

\section{Introduction}\label{Introduction}

Humans possess a remarkable ability to interact with the world despite slow neural transmission and low muscle bandwidth -- the long-standing paradox of human motor performance\cite{Hogan_2017}. This slow ``wet-ware'' results in feedback delays on the order of 25--350 ms. Human motor control must therefore rely on feedforward control. A kinematic regularity suggests an underlying structure: voluntary movements can be described as the temporal superposition of discrete motor primitives called submovements, each contributing a smooth, bell-shaped speed profile of finite duration \cite{woodworth1899, flash1985, hogan1984, plamondon1995kinematic, plamondon_2000, hogan2012, Giszter_2015}. These units are not an artefact of injury or task. The earliest movements of patients recovering from stroke are visibly fragmented, broken into short, highly stereotyped segments rather than smooth, continuous motion \cite{krebs1999}, yet the same units appear across more than a century of observation, from slow finger movements and eye saccades to cyclical movements, ballistic reaching, and the developing reaches of infants \cite{woodworth1899, Elliott_2001, vallbo1993, Crossman_1983, Collewijn_1988, Morasso_1981, Doeringer_1998a, park2017, vonHofsten_1991, Berthier_1996}, and indirectly underlying contact task \cite{Hermus_2020, Hermus_2024}. Submovements have formed the basis of numerous descriptive models of human actions \cite{Meyer_1988, Burdet_1998, Markkula_2018}. Submovement decomposition is, in addition, a sensitive probe of neuromotor recovery: submovement count, temporal overlap, and smoothness track change after stroke more finely than endpoint outcome measures \cite{krebs1999, rohrer2002, hogan2006}.

In parallel, the robotics community has developed a rich computational toolkit of parameterised movement primitives --- dynamical movement primitives, probabilistic movement primitives, Gaussian mixture models and Riemannian manifold formulations~\cite{ijspeert2013dynamical, paraschos2013probabilistic, calinon2016tutorial, calinon2020gaussians,Ficuciello18}. These tools have unlocked advances in motion planning, imitation learning, and skill transfer across embodiments. However, the primitives themselves are typically chosen to fit demonstrated trajectories or task constraints, not derived from human physiology. Submovement-Identifiable Decomposition (Sub-ID) bridges this gap by determining if and when the underlying onset time, duration, and amplitude of submovements can be recovered from human reaching data. This work therefore has the potential to unlock this powerful class of computational methods for human motor-control research. 

Recovering submovements from a kinematic recording is a hard inverse problem. Temporal superposition masks the individual primitives, as overlapping submovements sum into a single smooth, multi-peaked, or asymmetric velocity profile whose components cannot be separated by inspection. Here, we would like to group prior methods into three categories, distinguished by what each does to the problem to make it solvable. The first solves the decomposition as posed. Branch-and-bound \cite{rohrer2003} and scattershot \cite{rohrer2006} search for the globally optimal set of submovements, but their cost grows exponentially and becomes intractable beyond roughly ten primitives. The second gains tractability by constraining the problem. Search-space-reduction heuristics \cite{gowda2015}, generative priors, and fixed primitive libraries \cite{williams2006, williams2007modelling, raket2016separating} scale to longer signals, but introduce assumptions whose effect on the recovered submovements is never measured. The third learns the decomposition from data. SSSUMO \cite{rudakov2025} is trained on labels produced by the earlier algorithms and therefore inherits their biases while fixing the sampling rate and spatial dimension. Every scalable method trades exactness for assumptions, and none can reveal whether those assumptions have corrupted the result. The submovements returned may reflect the true signal, or they may reflect predominantly the bias introduced to make decomposition tractable.

This leads to a fundamental question: when is decomposition mathematically possible at all? When two primitives are separated by less than a minimum onset interval, their basis functions become nearly collinear, and the inverse problem loses its unique solution. This resolution limit has been acknowledged empirically\cite{rohrer2003, rohrer2006} but never formally characterised. The Cram\'er--Rao lower bound for amplitude estimation diverges as the Gram matrix becomes ill-conditioned, $\kappa(G) \rightarrow \infty$ --- the regime of information-theoretic impossibility \cite{candes2013, liu2021}. Without a criterion for identifiability, algorithm failure cannot be distinguished from mathematical impossibility: a spurious decomposition can match the data to within a fraction of a percent of residual error with no detectable signal of failure \cite{rohrer2003}.

\begin{figure}[ht]
    \centering
\includegraphics[width=1.0\linewidth]{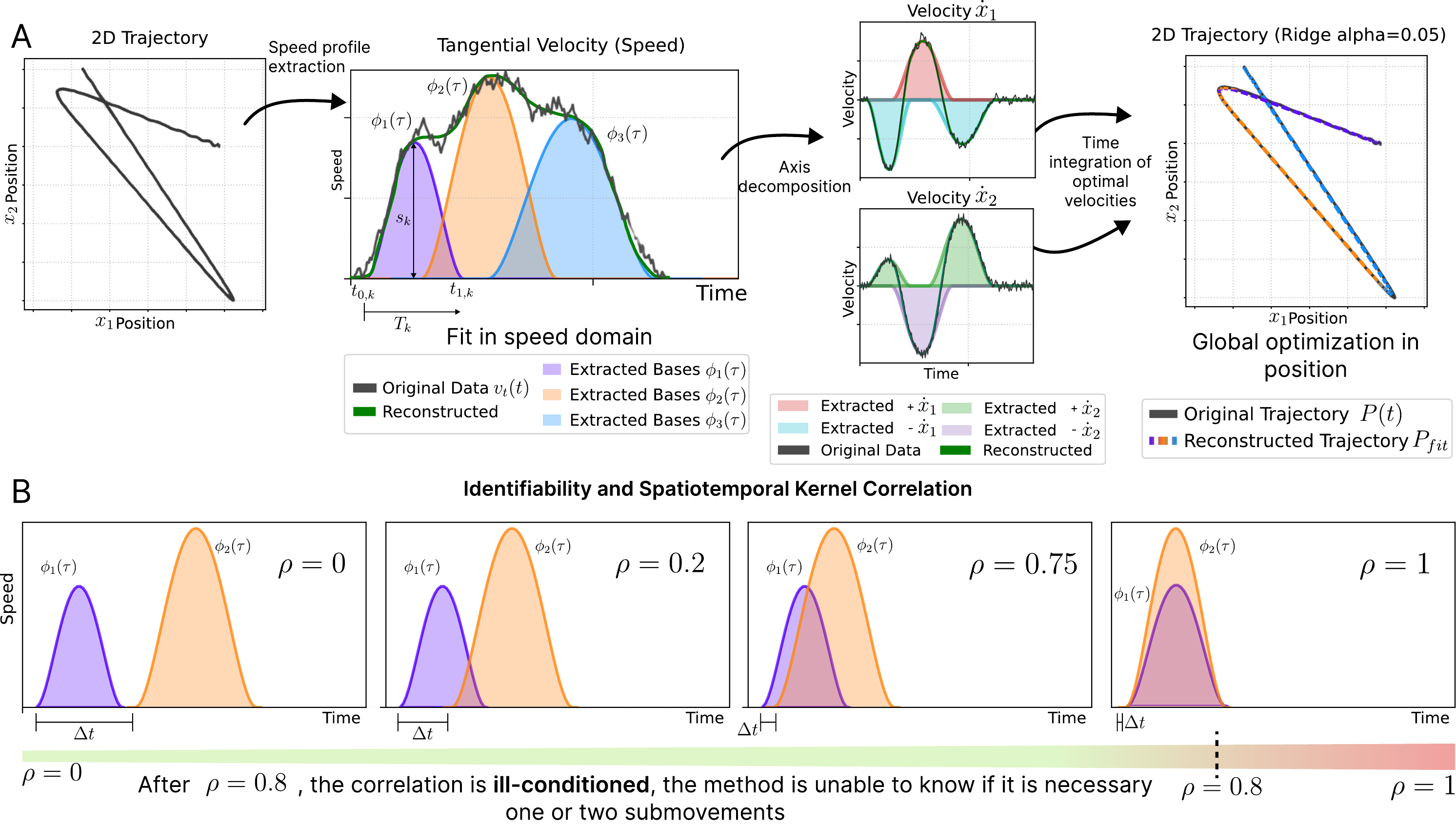}
    \caption{\textbf{Sub-ID framework and identifiability criterion.} 
\textbf{(A)} Submovement decomposition. The spatial trajectory is decoupled into scalar velocity to extract temporal velocities bases. These are spatially projected, integrated to position and scaled via Ridge Regression to prevent collinearity. A final global optimization minimizes positional error to yield the position reconstruction. 
\textbf{(B)} Spatiotemporal Kernel Correlation. Severe temporal overlap ($\rho \to 1$) causes unresolvable aliasing. Sub-ID mitigates this by incorporating spatial alignment, ensuring highly overlapped primitives remain distinguishable as long as their spatial directions diverge.}
    \label{fig:IntroScheme}
\end{figure}

We supply that criterion with the spatiotemporal kernel correlation $\rho_{i,j} = \cos(\psi_{i,j}) \cdot \rho(\Delta t)$, the off-diagonal entry of the Gram matrix above. Derived from the Cram\'er--Rao bound, it formalises when a pair of primitives is distinguishable, and shows that spatial divergence extends amplitude identifiability beyond purely temporal limits --- while onset detection in the scalar speed trace remains governed by the temporal factor alone. Sub-ID, the decomposition method built on this criterion, constrains the problem for tractability through heuristic initialisation and adaptive-ridge regularisation that biases the optimiser toward low-$\rho_{i,j}$ solutions.  The key idea of this work is to deliberately introduce bias into the algorithm to make it computationally tractable, then systematically show that, under the conditions we tested, this bias does not compromise the scientific value of the output. To do this, we demonstrate that Sub-ID recovers the ground-truth parameter distributions at primitive counts where exact methods are intractable. Furthermore, when tested on data statistically ill-posed for decomposition, the kernel correlation is a clear measure for identifying when the decomposition is not interpretable. These contributions allow the method to extract submovements from real three-dimensional, long-horizon movements --- a regime where existing methods drift or over-segment. Fig.~\ref{fig:IntroScheme} presents a visual and general scheme of the method Sub-ID.

\section{Results}
\subsection{Parameter recovery on synthetic data (unimodal) distributions}
\label{subsec:synthetic_results}

We first evaluate Sub-ID on synthetic trajectories, where the ground truth (GT) is known and the decomposition can be tested directly. Synthetic trajectories were generated at three primitive counts ($K = 3$, $8$, and $15$, with 100 unimodal trials per level), and Sub-ID was compared against search-space reduction via Gowda~\cite{gowda2015}, Scattershot~\cite{rohrer2006}, implementation, and SSSUMO~\cite{rudakov2025}to quantify its ability to decouple overlapping primitives and avoid spurious decompositions prior to testing on human data. Representative decompositions at different levels are presented in Fig.~1 of Supplementary Material S1.3. The reconstructed speed profiles exhibit a close agreement with the ground-truth data for different $K$ levels.

\begin{figure}[ht]
    \centering
    \includegraphics[width=1.0\linewidth]{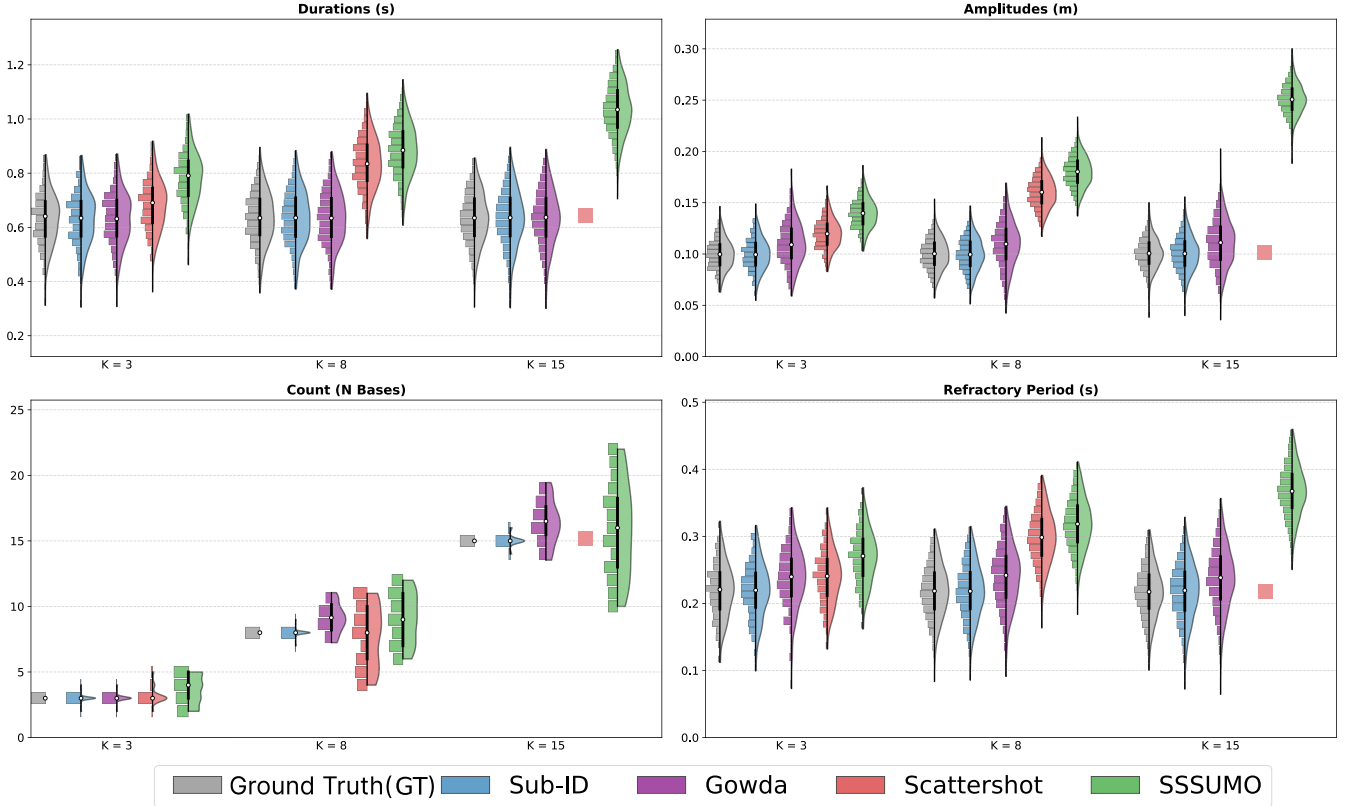}
    \caption{Comparison of kinematic parameter recovery against the GT for ($K=3, 8, 15$). Columns represent, from left to right: basis function durations, amplitudes, the number of extracted functions, and the refractory period. Results are shown for our method, Sub-ID (blue), SSSUMO (green), Scattershot (red) and Gowda (purple), against the original GT distribution (grey). The square symbol indicates that no valid solution was found by the algorithm for that specific scenario. Sub-ID consistently replicates the original distributions across all scenarios, whereas baseline methods exhibit significant parameter shifts, incorrect basis counts, or convergence failure as temporal density increases.}
    \label{fig:Comps}
\end{figure}
To evaluate reconstruction performance, we must first address the fundamentally different nature of the methods' termination conditions. Sub-ID, Gowda and Scattershot are optimization-based frameworks that explicitly target a termination threshold where the residual velocity error falls below $\tau_{tol} \approx 2\%$. By construction, Sub-ID achieves $\le 2\%$ velocity RMSE across all complexities, while Scattershot targets this same bound but is constrained by its maximum basis count when convergence fails. SSSUMO, by contrast, is a direct-inference neural network with no per-trial optimization loop; its velocity and position RMSE are simply whatever the trained network outputs. Because of these architectural disparities, we report velocity and position RMSE primarily as a baseline reference rather than a substantive comparison of success. All four methods fit primarily in the tangential velocity domain, where the bell-shaped primitives are cleanly defined. However, because small velocity residuals integrate and accumulate into position-domain drift over long sequences, reporting position RMSE serves as a necessary reference to certify that the velocity fits do not silently diverge. For instance, Sub-ID's global optimization in the position domain maintains a position RMSE of $0.012,\mathrm{m}$ even at $K=15$. Scattershot reaches its maximum number of basis functions when it is unable to satisfy the $2\%$ threshold, successfully converging for $K=3$, degrading at $K=8$, and failing at $K=15$. Gowda's method produces accurate results in the velocity domain; however, errors accumulate when converting the solution to the position domain, yielding a minimum position error of $0.015,\mathrm{m}$ for $K=3$, which progressively increases with the number of submovements, reaching approximately $0.06,\mathrm{m}$ at $K=15$. In contrast, the unoptimized position error of SSSUMO reaches $0.120,\mathrm{m}$ at $K=15$.

With these reconstruction baselines established as a convergence reference, the truly informative axis for studying human motor control is parameter recovery. The core scientific question is whether each method can accurately replicate the underlying ground-truth (GT) latent distributions of submovement duration, amplitude, basis count, and refractory period (Fig.~\ref{fig:Comps}). A direct comparison with the GT highlights stark differences in robustness among the evaluated methods. Sub-ID replicates the GT parameter distributions with high fidelity across all four kinematic criteria, maintaining its accuracy at every level of complexity. The baseline methods, by contrast, exhibit distinct and progressive failure modes. SSSUMO captures the rough shape of each distribution but deviates from the exact GT values even in the simplest scenario ($K=3$), and systematically overestimates the required primitive count. Scattershot identifies the correct number of bases at $K=3$, albeit with deviations in the remaining parameters, but loses this capability at $K=8$. At $K=15$, it completely fails to converge within the maximum limit of 30 basis functions. Among the compared approaches, Gowda's method yields the closest performance to Sub-ID, identifying the mean of each distribution without substantial deviation for the duration, amplitude, and refractory period. Nevertheless, it shows a tendency to overfit in the $K=8$ and $K=15$ scenarios, finding more basis functions than those contained in the GT decomposition.

The kernel correlation of the recovered decomposition is itself a self-diagnostic. The synthetic ground truth was generated with well-separated primitives, so its pairwise $\rho_{i,j}$ is low by construction; when a method returns a decomposition whose $\rho_{i,j}$ approaches 1, the recovered bases are nearly collinear, the Gram matrix is singular, and the solution lies in its null space --- a family of collinear combinations that fit the data equally well, none of which can be told apart by the data alone. By this criterion, Sub-ID tracks the ground truth: even at $K=15$, critical overlaps ($>0.8$) barely reach $3.00\%$, and the maximum correlation peaks at $\rho = 0.45$ --- well clear of the singular regime. Gowda offers a robust baseline that performs accurately at lower complexities ($\rho = 0.32$ at $K=3$), but it increasingly struggles to maintain independence in dense trajectories; by $K=15$, its correlation rises to $\rho = 0.68$ with $12.30\%$ of overlaps crossing the critical threshold, indicating a partial loss of identifiability. SSSUMO, drifts progressively into the null space --- $\rho$ rises from $0.5\text{--}0.6$ at $K=3$ to $0.96$ at $K=15$, with $51.83\%$ of its basis functions in extreme overlap. Scattershot is well-behaved at $K=3$ but interferes substantially at $K=8$, and fails to converge at $K=15$. The gap between the low $\rho$ of the generating process and the high $\rho$ of the SSSUMO output is, on its own, evidence that the decomposition is spurious.
%

\begin{figure}[H]
    \centering
\includegraphics[width=0.92\linewidth]{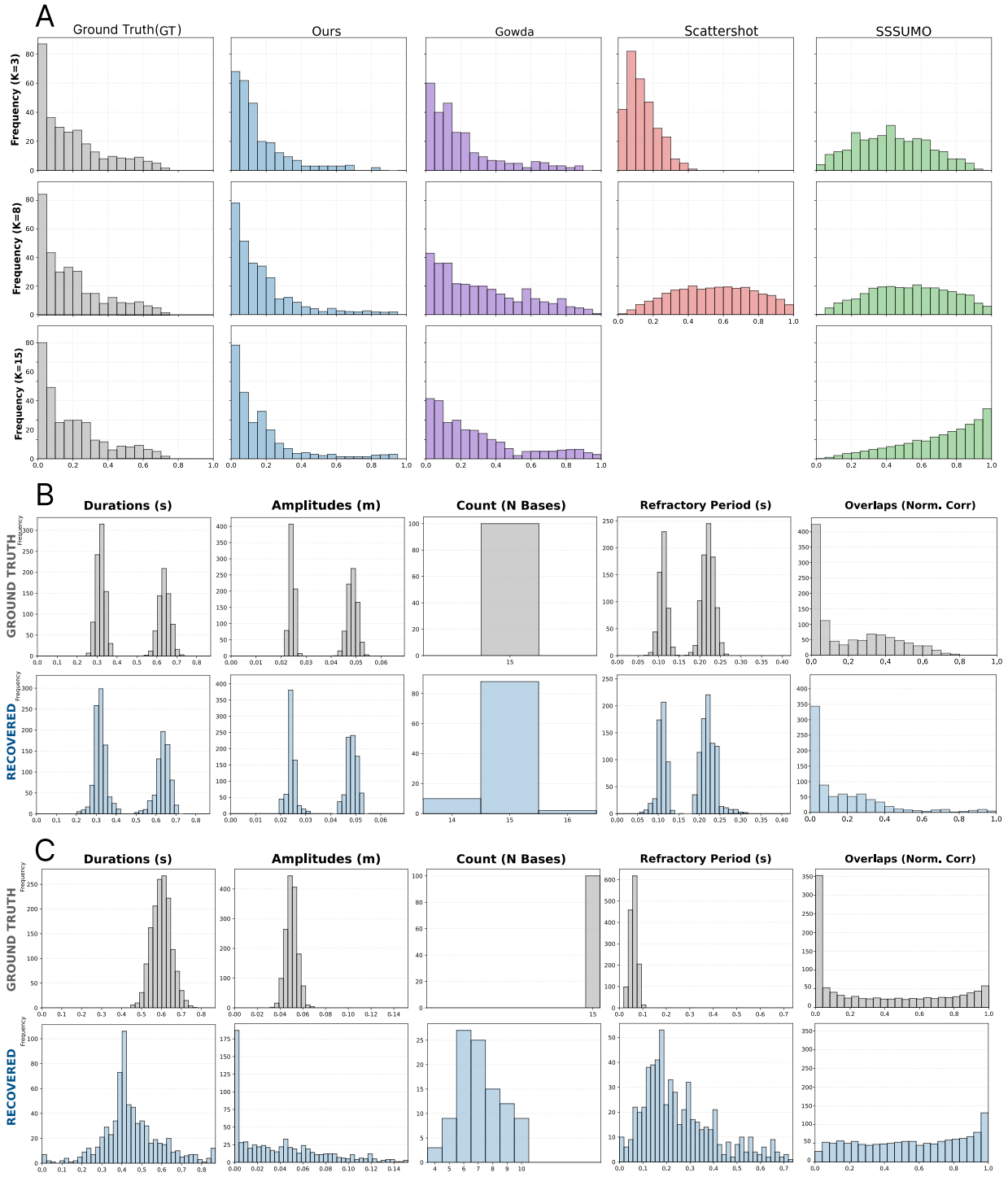}
    \caption{Evaluation of extracted kinematic parameters and kernel overlaps. \textbf{(A)} Overlap distributions for varying complexities ($K=\{3, 8, 15\}$). Sub-ID consistently matches the GT low-overlap profiles. Gowda offers a resilient baseline but exhibits a moderate increase in overlapping bases at $K=15$. SSSUMO exhibits increasing correlation, culminating in heavy overlap near 1.0 at $K=15$. Scattershot suffers interference and fails to converge at $K=15$. \textbf{(B)} Bimodal GT comparison ($K=15$). Sub-ID robustly captures the underlying dual structure and maintains bounded overlaps, despite minor variance in the recovered basis count (14--16 vs. 15). \textbf{(C)} Unimodal GT extreme overlap scenario ($K=15$). Severe temporal aliasing exposes identifiability limits: closely spaced submovements are merged, causing a significant underestimation of the basis count and a drastic increase in overlap. For \textbf{(B)} and \textbf{(C)}, blue presents our method and gray the Ground Truth (GT).}
    \label{fig:ResultsTogether}
\end{figure}

Fig.~\ref{fig:ResultsTogether}A presents in a visual way the underlying behavior of each method concerning temporal aliasing. Setting a 0.8 redundancy threshold, our algorithm closely matches the GT overlapping distributions in all scenarios. By consistently maintaining low overlap indices, it effectively prevents the creation of redundant primitives (Table~\ref{tab:combined_metrics_all} (Part A)). Gowda provides a resilient heuristic baseline, avoiding severe interference at lower complexities, yet its tendency to overfit dense trajectories introduces moderate temporal aliasing, with $12.30\%$ of its bases crossing the redundancy threshold by $K=15$. In contrast, while Scattershot maintains low overlap at $K=3$, it displayed significant basis interference at $K=8$. SSSUMO struggles even earlier, showing medium-high correlations ($0.5\text{--}0.6$) at $K=3$. As complexity increases, this overlap severely degrades, culminating at $K=15$ where most SSSUMO basis functions exhibit overlaps near $1.0$, resulting in spurious decompositions. Given the synthetic nature of this benchmark, arbitrary increases in the number of simulated trials would render standard statistical significance tests ($p$-values) uninformative, as even practically irrelevant differences would appear statistically significant. Therefore, our comparative evaluation focuses on the magnitude of the errors (e.g., RMSE) and the extent of distributional divergence, which better reflect the practical utility of the decomposition. Table~\ref{tab:combined_metrics_all} (Part A) quantitatively evaluates the methods across different metrics.
\newpage

\subsection{Generalisation to heterogeneous (bimodal) distributions}
Following the success in unimodal scenarios, we evaluated Sub-ID in a more restrictive setting: reconstructing tasks with parameters drawn from two distinct bimodal distributions at the limiting case of $K=15$. Across 100 examples (Fig.~\ref{fig:ResultsTogether}B), the method successfully identifies the dual nature of the signal, capturing the clear separation in durations, amplitudes, and refractory periods.

Unlike the unimodal case, the recovered distributions exhibit slight additional dispersion and small tails in the Gaussian profiles. However, these modes never merge, thereby preserving the bimodal independence of the synthetic data. The estimated number of basis functions ($K$) shows minor variance, occasionally oscillating between 14 and 16 primitives rather than exactly 15. Regarding overlapping, the algorithm generally maintains a distribution similar to the GT, with only rare instances exceeding the $0.8$ threshold limit. The quantitative evaluation of the bimodal distribution reconstruction results is detailed in Supplementary Material S4.1.

\subsection{Graceful failure in the non-identifiable regime}\label{sect:garcefulFailure}

We next probe Sub-ID in a regime where decomposition is mathematically ill-posed: a unimodal synthetic scenario ($K=15$) constructed with extreme primitive overlap. When successive primitive onsets fall below the resolution limit $\Delta t_{\min}(\tau)$, basis functions become nearly linearly dependent. No algorithm can uniquely recover the contributing primitives from kinematics alone, and prior fast methods fail silently in this regime: false decompositions can match the data to within $0.5\%$ residual error \cite{rohrer2003}, with no signal distinguishing them from valid solutions. Sub-ID fails differently (see Fig.~\ref{fig:ResultsTogether}C). The adaptive ridge penalty grows with the conditioning of the Gram matrix favoring low-$\rho_{i,j}$ solutions (Sect.~\ref{StepbyStep}). A high $\rho_{i,j}$ in the recovered decomposition is therefore not a chance event. It indicates that the optimiser was forced into the ill-conditioned regime because no low-correlation solution fits the data. In our controlled synthetic trials, as the onset latency was reduced, $\rho_{i,j}$ rose toward $1$ and Sub-ID returned a lower-$K$ solution rather than a spurious full-$K$ decomposition. The failure was surfaced, not hidden (Supplementary Material S4.2).


\subsection{Real Data: 2D long and short horizon tasks}

Following synthetic validation, we extended our empirical evaluation to real tasks extracted from human subjects using two datasets at opposite ends of the complexity spectrum: \textbf{Letters}~\cite{williams2006, williams2007modelling}, for short low-density strokes, and \textbf{PushT}, for continuous manipulation tasks with long sequences. In these cases, as no Ground Truth is available for empirical data, the exact solution cannot be determined with absolute certainty. 

\begin{figure}[ht]
    \centering
    \includegraphics[width=1.0\linewidth]{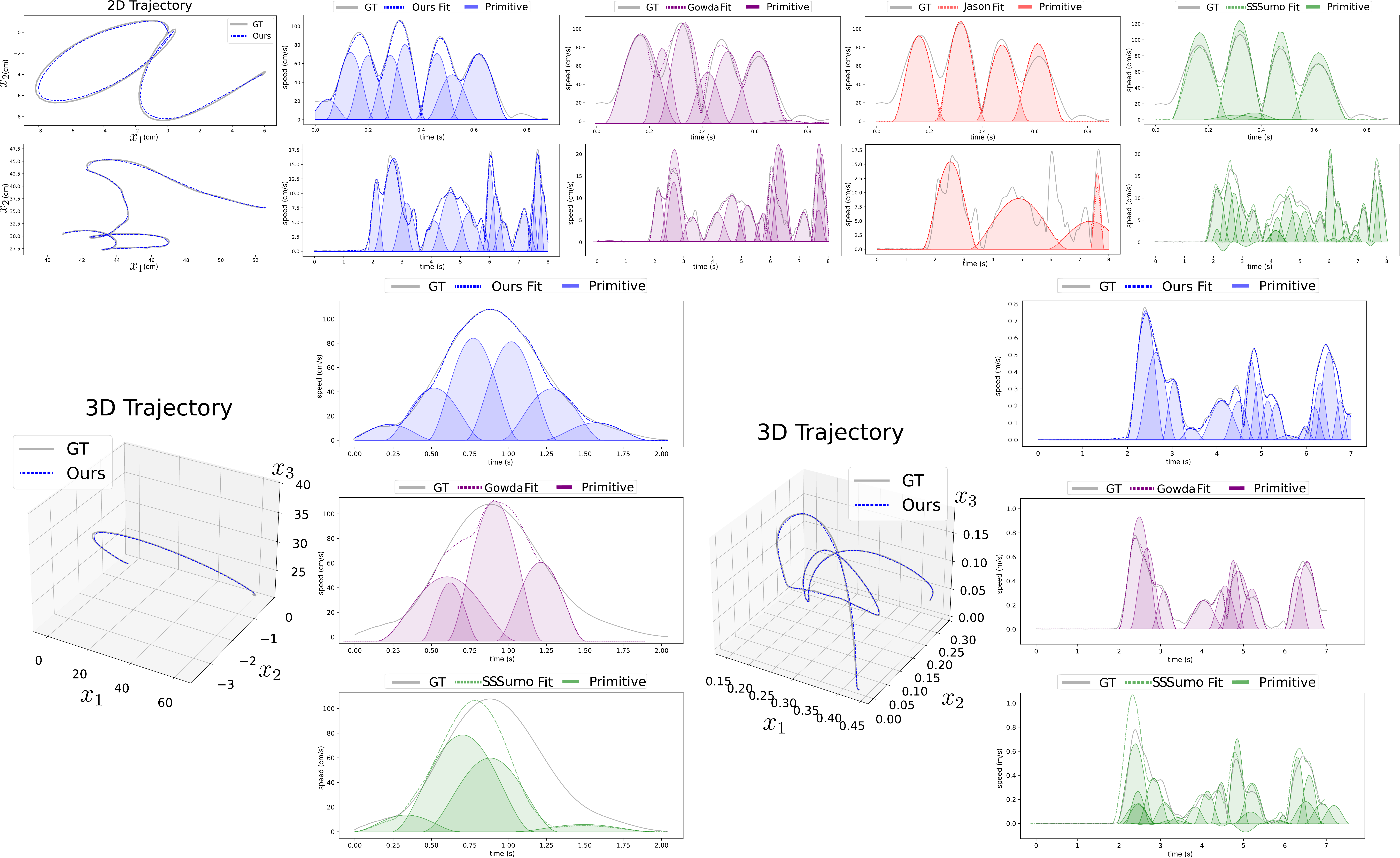}
    \caption{Qualitative evaluation of kinematic decompositions in 2/3D tasks. The figure presents spatial trajectory reconstructions alongside their tangential velocity profiles and extracted primitives. For the 2D tasks, the top row shows a simple stroke from the Letters dataset ('A'), and the bottom row a continuous sequence from the PushT dataset. For the 3D tasks, the top row displays a short trajectory from the Moving dataset, and the bottom row a continuous sequence from the PushT 3D dataset. Across both domains, Sub-ID (blue) fits the signal using a concise set of primitives. In contrast, SSSUMO (green) exhibits overfitting in all scenarios, fragmenting the movement into an excess of highly overlapped primitives, clearly observable in the PushT sequences. Scattershot (red) is included in the 2D evaluation—where it shows underfitting in long tasks by omitting secondary peaks and valleys—but is excluded from the 3D tasks due to its difficulty scaling to high-dimensional and complex sequences.}
    \label{fig:2_3DExperiment}
\end{figure}

Fig.~\ref{fig:2_3DExperiment} provides a qualitative evaluation where it is observed that our method traces the original spatial trajectory without deviations and captures the signal morphology by fitting velocity peaks with well-defined primitives without falling into overfitting. In contrast, the SSSUMO method incurs severe overfitting in both simple tasks and PushT, fragmenting the movement into an excessive number of compressed basis functions that generate a noisy profile. Scattershot shows evident underfitting in long-horizon tasks, being unable to generate enough primitives to reconstruct secondary velocity valleys and peaks. Finally, the Gowda algorithm strikes a middle ground: it successfully avoids catastrophic underfitting but still introduces a moderately noisy profile, relying on wider and occasionally redundant basis functions as trajectory complexity increases.

Table~\ref{tab:combined_metrics_all} (Part B) presents the quantitative metrics. In terms of kinematic fidelity, Sub-ID provides a reasonable and consistent reconstruction, maintaining millimetric precision ($0.1\text{ cm}$ for Letters; $0.14\text{ cm}$ for PushT) and keeping the relative position error exceptionally low ($\le 1.87\%$) by operating in both velocity and position domains, while explaining most of the kinematic variance ($\text{Vel } R^2 > 0.96$). As anticipated from the simulated data results, the baseline methods face challenges when scaling to longer continuous records. SSSUMO exhibits higher relative positional deviations that scale from $7.98\%$ in short strokes to $23.35\%$ in extended manipulation. Meanwhile, Scattershot structurally encounters difficulties maintaining its fit on long sequences like PushT, where its lack of convergence drives the relative position error to $57.35\%$ and causes a severe drop in kinematic variance explanation ($\text{Vel } R^2 = 0.1808$; $\text{RMSE} = 40.03\text{ m/s}$). Gowda offers a more stable baseline than Scattershot but shows noticeable degradation on extended sequences; its relative position error increases from $2.45\%$ in short strokes (Letters) to $8.54\%$ and $10.54\%$ in the 2D and 3D PushT datasets, respectively, accompanied by a moderate drop in velocity variance explanation ($\text{Vel } R^2 \approx 0.81-0.85$).

Beyond kinematic reconstruction, structural identifiability metrics reveal critical differences in how the algorithms decompose the movement. In the complex PushT dataset, Sub-ID extracts an average of $K=14.62$ primitives, maintaining a highly structured temporal sequence with a mean overlap of just $0.0549$ and keeping critical kernel correlations (overlaps $>0.8$) to a minimal $2.17\%$. This ensures that each primitive resolves a distinct kinematic event without mathematical ambiguity. SSSUMO, by contrast, illustrates a total loss of identifiability by extracting approximately twice the primitives ($K=49.8$), leading to a maximum overlap of $1.0$. This perfect collinearity between basis functions indicates extreme structural redundancy, reflecting the ill-conditioning of methods lacking strict spatiotemporal constraints. Meanwhile, Scattershot struggles computationally with these extended tasks, failing to reach success thresholds entirely and producing sparse, non-convergent solutions. Gowda manages to avoid the complete structural collapse seen in SSSUMO but still succumbs to partial aliasing; in the 2D PushT task, it extracts an over-segmented $K=18.21$ primitives with $12.45\%$ of them crossing the critical overlap threshold (rising to $16.45\%$ in 3D PushT), revealing an increasing reliance on collinear, redundant bases to maintain its fit.

\subsection{Real Data: 3D long and short horizon tasks}
Finally, we evaluate algorithmic scalability in the 3D domain using the \textbf{Moving}~\cite{grimme2012naturalistic} and \textbf{PushT 3D} datasets. The Scattershot algorithm was excluded from this phase, as its iterative architecture fails to converge on long tasks and degrades severely with increased dimensionality. Consequently, the analysis focuses on a comparison between Sub-ID, SSSUMO and Gowda. 

Qualitative results (Fig.~\ref{fig:2_3DExperiment}) show that Sub-ID (blue) extracts a coherent, ordered set of submovements that naturally explain velocity peaks in both 3D tasks. Gowda's algorithm (purple) provides a visually plausible fit but begins to exhibit signs of temporal aliasing, relying on clustered and overlapping primitives to reproduce the 3D profile. Moreover, the resulting aliasing occasionally leads to gaps in the reconstruction, leaving portions of the signal insufficiently represented by the recovered basis functions. In contrast, SSSUMO (green) collapses under the added dimensionality, injecting a chaotic mass of overlapping primitives—a failure particularly evident in the PushT 3D profile.

Table~\ref{tab:combined_metrics_all} (Part B) presents the metrics. In terms of kinematic fidelity, Sub-ID provides a reasonable and consistent reconstruction, maintaining near-zero absolute position errors ($0.13\text{ cm}$ and $0.17\text{ cm}$) while keeping the relative position error exceptionally low and bounded ($0.95\%$ for Moving; $2.12\%$ for PushT 3D). Furthermore, it explains most of the empirical variance ($\text{Vel } R^2 > 0.93$) with minimal velocity RMSE ($0.0048$ and $0.0139\text{ m/s}$). As anticipated from the simulated data results, the baseline method faces scaling challenges over these higher-dimensional continuous records. Gowda demonstrates moderate degradation when scaling to 3D; its relative position error increases from $3.25\%$ in the Moving task to $10.54\%$ in PushT 3D, alongside a drop in velocity variance explanation ($\text{Vel } R^2$ falling from $0.8845$ to $0.8145$). SSSUMO exhibits higher relative positional deviations that scale from $14.22\%$ in the Moving task up to $28.64\%$ in PushT 3D (with absolute errors reaching $3.63\text{ m}$), while failing to exceed a $\text{Vel } R^2$ of $0.8527$ and showing an increased velocity RMSE ($3.4870\text{ m/s}$). These limitations stem from using only scalar velocity data, which leads to reconstruction deviations without full directional information.
\begin{table*}[htbp]
    \centering
    \caption{Performance comparison across synthetic tasks of varying complexity (top) and Comparative Evaluation on 2D and 3D Datasets (bottom).}
    \label{tab:combined_metrics_all}

    \textbf{Part A: Synthetic tasks of varying complexity ($K=3, 8, 15$)}
    \vspace{1ex}
    
    \resizebox{\textwidth}{!}{%
        \begin{tabular}{l ccc ccc ccc ccc ccc}
        \toprule
        & \multicolumn{3}{c}{Overlaps $> 0.8$ ($\mu \pm \sigma$)}
        & \multicolumn{3}{c}{Position RMSE (cm)}
        & \multicolumn{3}{c}{Velocity RMSE (m/s)}
        & \multicolumn{3}{c}{Co-lineality $\Phi$}
        & \multicolumn{3}{c}{Spatiotemporal $\rho$} \\
        \cmidrule(lr){2-4} \cmidrule(lr){5-7} \cmidrule(lr){8-10} \cmidrule(lr){11-13} \cmidrule(lr){14-16}
        \textbf{Method} & K=3 & K=8 & K=15 & K=3 & K=8 & K=15 & K=3 & K=8 & K=15 & K=3 & K=8 & K=15 & K=3 & K=8 & K=15 \\
        \midrule
        \textbf{GT} & 0.00$\pm$0.00\% & 0.00$\pm$0.00\% & 0.00$\pm$0.00\% & 0.0 & 0.0 & 0.0 & 0.000 & 0.000 & 0.000 & 1.0 & 1.2 & 1.5 & -- & -- & -- \\
        \addlinespace
        \textbf{Sub-ID} & 0.84$\pm$0.19\% & \textbf{2.08$\pm$1.06\%} & \textbf{3.00$\pm$1.35\%} & \textbf{0.2} & \textbf{0.4} & \textbf{1.2} & \textbf{0.005} & \textbf{0.007} & \textbf{0.009} & \textbf{1.5} & \textbf{2.8} & \textbf{8.2} & 0.32 & \textbf{0.35} & \textbf{0.45} \\
        \addlinespace
        \textbf{Gowda} & 0.95$\pm$0.25\% & 7.20$\pm$2.15\% & 12.30$\pm$3.20\% & 0.4 & 1.5 & 6.0 & 0.006 & 0.018 & 0.035 & 3.5 & 11.0 & 27.3 & 0.32 & 0.45 & 0.68 \\
        \addlinespace
        \textbf{Scattershot} & \textbf{0.00$\pm$0.00\%} & 17.14$\pm$6.98\% & -- & 0.5 & 8.5 & -- & \textbf{0.005} & 0.057 & -- & 4.0 & 150.0 & -- & \textbf{0.30} & 0.82 & -- \\
        \addlinespace
        \textbf{SSSUMO} & 4.67$\pm$13.35\% & 15.46$\pm$6.31\% & 51.83$\pm$4.61\% & 1.5 & 4.5 & 12.0 & 0.012 & 0.073 & 1.673 & 12.0 & 85.0 & 450.0 & 0.72 & 0.88 & 0.96 \\
        \bottomrule
        \end{tabular}
    }
    
    \vspace{2ex}
    
    \textbf{Part B: Comparative Evaluation on 2D and 3D Datasets}
    \vspace{1ex}
    
    \resizebox{\textwidth}{!}{
        \begin{tabular}{llccccccccc}
        \toprule
        \textbf{Dataset} & \textbf{Method} & \textbf{K} & \textbf{Rate (K/s)} & \textbf{Pos Error (cm)} & \textbf{Pos Error (\%)} & \textbf{Vel $R^2$} & \textbf{Vel RMSE (m/s)} & \textbf{Mean Overlap} & \textbf{Max Overlap} & \textbf{Overlaps (\%) $> 0.8$} \\
        
        \midrule
        \multicolumn{11}{c}{\textbf{2D Datasets}} \\
        \midrule
        
        \multirow{4}{*}{Letters} & Scattershot & 3.25 $\pm$ 1.02 & 3.9532 $\pm$ 1.2065 & 17.95 $\pm$ 7.52 & 4.77 $\pm$ 1.32 & 0.7791 $\pm$ 0.2173 & 0.0462 $\pm$ 0.0825 & 0.4644 $\pm$ 0.1553 & 0.7330 $\pm$ 0.2511 & 3.6342 $\pm$ 2.8464 \\
         & Sub-ID & 6.25 $\pm$ 0.68 & 2.4576 $\pm$ 0.7285 & \textbf{0.10 $\pm$ 0.01} & \textbf{0.91 $\pm$ 0.49} & \textbf{0.9633 $\pm$ 0.0449} & \textbf{0.0052 $\pm$ 0.0002} & \textbf{0.1497 $\pm$ 0.0683} & \textbf{0.7232 $\pm$ 0.0930} & \textbf{1.7354 $\pm$ 0.8862} \\
         & SSSUMO & 8.06 $\pm$ 1.39 & 5.9475 $\pm$ 2.0511 & 26.88 $\pm$ 9.46 & 7.98 $\pm$ 2.67 & 0.9052 $\pm$ 0.0068 & 1.6665 $\pm$ 0.9579 & 0.2541 $\pm$ 0.0519 & 0.9781 $\pm$ 0.0706 & 8.0100 $\pm$ 3.5317 \\
         & Gowda & 7.51 $\pm$ 1.20 & 2.9510 $\pm$ 0.8120 & 0.85 $\pm$ 0.21 & 2.45 $\pm$ 0.85 & 0.9214 $\pm$ 0.0512 & 0.0156 $\pm$ 0.0042 & 0.2014 $\pm$ 0.0812 & 0.8412 $\pm$ 0.0815 & 5.1245 $\pm$ 2.1054 \\
        \midrule
        \multirow{4}{*}{PushT} & Scattershot & 6.29 $\pm$ 0.86 & 0.7867 $\pm$ 0.1073 & 28.48 $\pm$ 13.04 & 57.35 $\pm$ 12.05 & 0.1808 $\pm$ 0.1868 & 40.0352 $\pm$ 12.0398 & --- & --- & --- \\
         & Sub-ID & 14.62 $\pm$ 2.02 & 1.8275 $\pm$ 0.2529 & \textbf{0.14 $\pm$ 0.05} & \textbf{1.87 $\pm$ 0.28} & \textbf{0.9845 $\pm$ 0.0170} & \textbf{0.0124 $\pm$ 0.0089} & \textbf{0.0549 $\pm$ 0.0181} & \textbf{0.8672 $\pm$ 0.0168} & \textbf{2.1766 $\pm$ 1.3860} \\
         & SSSUMO & 27.31 $\pm$ 3.49 & 3.4137 $\pm$ 0.4366 & 47.47 $\pm$ 14.41 & 23.35 $\pm$ 6.12 & 0.7745 $\pm$ 0.1240 & 0.5511 $\pm$ 0.2948 & 0.1420 $\pm$ 0.0598 & 1.0000 $\pm$ 0.0000 & 5.5906 $\pm$ 1.5141 \\
         & Gowda & 18.21 $\pm$ 2.45 & 2.2750 $\pm$ 0.3105 & 1.85 $\pm$ 0.84 & 8.54 $\pm$ 2.15 & 0.8542 $\pm$ 0.1245 & 0.0451 $\pm$ 0.0152 & 0.1512 $\pm$ 0.0415 & 0.9514 $\pm$ 0.0210 & 12.4510 $\pm$ 4.2150 \\
        
        \midrule
        \multicolumn{11}{c}{\textbf{3D Datasets}} \\
        \midrule
        
        \multirow{3}{*}{Moving} & Sub-ID & 6.09 $\pm$ 1.10 & 5.6557 $\pm$ 0.9218 & \textbf{0.13 $\pm$ 0.04} & \textbf{0.95 $\pm$ 0.32} & \textbf{0.9374 $\pm$ 0.0717} & \textbf{0.0048 $\pm$ 0.0004} & \textbf{0.2629 $\pm$ 0.0793} & \textbf{0.7200 $\pm$ 0.1166} & \textbf{2.4299 $\pm$ 0.9683} \\
         & SSSUMO & 5.38 $\pm$ 1.41 & 5.0134 $\pm$ 1.3805 & 363.87 $\pm$ 85.53 & 14.22 $\pm$ 3.85 & 0.8318 $\pm$ 0.0629 & 3.4870 $\pm$ 0.7741 & 0.3460 $\pm$ 0.1248 & 0.9181 $\pm$ 0.0831 & 27.5329 $\pm$ 8.8390 \\
         & Gowda & 7.21 $\pm$ 1.45 & 6.6540 $\pm$ 1.1540 & 0.65 $\pm$ 0.24 & 3.25 $\pm$ 1.14 & 0.8845 $\pm$ 0.0845 & 0.0125 $\pm$ 0.0035 & 0.3015 $\pm$ 0.0914 & 0.8814 $\pm$ 0.0754 & 8.2145 $\pm$ 3.1045 \\
        \midrule
        \multirow{3}{*}{PushT} & Sub-ID & 16.50 $\pm$ 1.19 & 2.0625 $\pm$ 0.1488 & \textbf{0.17 $\pm$ 0.05} & \textbf{2.12 $\pm$ 0.41} & \textbf{0.9560 $\pm$ 0.0382} & \textbf{0.0139 $\pm$ 0.0061} & \textbf{0.0520 $\pm$ 0.0119} & \textbf{0.8574 $\pm$ 0.0182} & \textbf{5.9787 $\pm$ 1.2127} \\
         & SSSUMO & 27.30 $\pm$ 3.98 & 3.4125 $\pm$ 0.4981 & 18.36 $\pm$ 5.70 & 28.64 $\pm$ 5.92 & 0.8527 $\pm$ 0.2561 & 0.2371 $\pm$ 0.0351 & 0.0920 $\pm$ 0.0797 & 1.0000 $\pm$ 0.0000 & 12.2570 $\pm$ 3.9858 \\
         & Gowda & 20.45 $\pm$ 3.11 & 2.5560 $\pm$ 0.4150 & 2.45 $\pm$ 1.05 & 10.54 $\pm$ 3.45 & 0.8145 $\pm$ 0.1542 & 0.0654 $\pm$ 0.0214 & 0.1814 $\pm$ 0.0612 & 0.9654 $\pm$ 0.0214 & 16.4510 $\pm$ 5.4510 \\
        
        \bottomrule
        \end{tabular}
    }
    
    \vspace{1ex} 
    {\raggedright \footnotesize \textit{Note:} Bold values denote the best performance in each metric for their respective evaluations. \par}
\end{table*}

Beyond basic kinematic reconstruction, structural identifiability metrics reveal critical differences in how the algorithms decompose the movement, especially as these issues peak in the PushT 3D task. Sub-ID extracts an average of $16.5$ basis functions ($2.04\text{ K/s}$) with a mean overlap of $0.0520$, keeping critical interactions (overlaps $>0.8$) tightly restricted to $5.97\%$. The added dimensionality heavily impacts Gowda's identifiability; it over-segments the PushT 3D trajectory into $K=20.45$ primitives, driving its critical interactions to $16.45\%$ and its maximum overlap to $0.9654$. Conversely, SSSUMO introduces a higher density of primitives ($43.7$ primitives; $3.25\text{ K/s}$), leading to an absolute maximum overlap of $1.0$ and causing $12.25\%$ of interactions to become critical. This confirms that without strict spatiotemporal constraints, SSSUMO's bases become collinear and redundant, reflecting the ill-conditioning of methods lacking spatial coupling when scaling to 3D tasks.

\section{Discussion}

The temporal aliasing failure mode identified here is an information-theoretic bound, not an algorithmic deficiency. When two submovements are placed too close together, they become observationally equivalent to a single primitive, and no algorithm operating on kinematics alone can resolve them without additional prior information. Previous work acknowledged this limitation empirically \cite{rohrer2003, rohrer2006} with the observation that false decompositions can fit the data to within 0.5\% residual error, leaving no signal that distinguishes valid solutions from spurious ones. The present result formalizes that observation and converts it into a computable diagnostic. The kernel correlation and the condition number of the Gram matrix together measure how well-posed a decomposition is: low values confirm the primitives are distinguishable, while values approaching the collinear limit signal information-theoretic impossibility, the regime in which the variance of any amplitude estimate diverges. The same conditioning principle governs spectral super-resolution \cite{candes2013, liu2021, li2022} and high-density EMG decomposition \cite{Klotz_2025}; submovement decomposition is, in this sense, one instance of a broader family of identifiability-limited inverse problems.

Sub-ID gains its scalability by constraining the decomposition problem. The Heuristic Peak-Based Detection restricts the search space, and the adaptive ridge biases the optimiser away from collinear primitive sets. These constraints are not free: they are biased by design, and they are the precise reason Sub-ID escapes the exponential cost of unconstrained global optimisation. The substantive test of any such method is whether the bias compromises the scientifically informative output --- the distribution of submovement parameters that underlies motor-control inference and robot learning from demonstration. Across $K = 3$, $8$, and $15$, this work quantified the extent to which Sub-ID can identify the underlying parameter distributions, and in the non-identifiable regime ($\rho_{i,j} \to 1$), the failure is observable.

The value of a decomposition lies in what it reveals about composition. Submovements have been studied for over a century, yet the question that motivates them is not only whether a movement can be broken into pieces, but how the nervous system assembles those pieces into purposeful action~\cite{woodworth1899, krebs1999}. That question is challenging to ask unless it is possible to decompose longer actions where numerous submovements are superimposed. Sub-ID raises the temporal ceiling of decompositions in this regime. This opens the potential to use submovement parameter information to study human actions with tools from robot imitation learning frameworks (e.g., DMP, ProMP, GMM, GPs)~\cite{saveriano2023dynamic,li2023prodmp,prados2024learning,Calinon19MM,Ficuciello18,prados2026learning}.


A deeper question underlies every submovement decomposition, including this one. The method assumes that the movement was composed of separable primitives in the first place, yet the nervous system is under no obligation to compose actions so that a later observer can pull them apart. Whether it does is exactly what the kernel correlation measures: on real data, $\rho_{i,j}$ reports how far a recorded movement sits from the singular regime where recovery becomes impossible. Our observations so far are encouraging recovered $\rho_{i,j}$ in the human datasets remain low, with only occasional pairs approaching the resolution threshold and no sign that unresolvable overlap is the norm. Whether this holds across tasks and populations is an open question, and it is the precondition for everything built here. If human movement is generically non-identifiable, no decomposition based on kinematics alone can avoid this. However, our simulation and experimental results with Sub-ID, presented here in, up to 15 submovements, suggest that the motor commands are identifiable in the task we investigated. Of course, we cannot rule out that other datasets carry additional structure such as patterned parameter distributions, or systematic relations between submovements. Further work will solidify the answer to this question. In these next steps, the key ideas presented here provide a clear path forward: use of the Sub-ID method, the approach of assessing identifiability, and the methodology of decomposing synthetic trajectories with similar properties to experimental data.  

\subsection{Limitations}
Sub-ID's identifiability diagnostic carries one important caveat on real data: $\rho_{i,j}$ depends on the recovered direction vectors, which are themselves estimated outputs of the decomposition. A spurious solution containing cancelling bases could in principle report low $\rho_{i,j}$ if its estimated directions happened to be orthogonal, providing false assurance that the decomposition is well-posed. Until estimated directions are validated against synthetic ground truth in real-data regimes, the $\rho_{i,j}$ readout should be treated as necessary but not sufficient evidence of identifiability. 

In addition, SSSUMO was evaluated as released; whether retraining on Sub-ID-generated labels would close part of the gap reported here remains an open question for fairness comparisons.

\subsection{Future Work}

One choice in Sub-ID remains open to future revision: the coordinate frame of the final fit. Humans do not appear to perceive or generate motion in Cartesian coordinates \cite{Hogan_1990b, Fasse_2000}, yet the position-domain refinement is carried out in Cartesian space. We adopt it not as a claim about the nervous system but for compatibility: it is the frame in which most robot learning from demonstration operates, and fitting there keeps the position error small enough for that use. Velocity transforms between coordinate systems through the Jacobian of the kinematic map, so the same objective could be expressed in object-centered coordinates without altering the earlier stages. Which frame and geometry is right for submovement decomposition remains an open question.

Sub-ID recovers submovement timing in one domain and amplitude in another, and the division follows from the structure of the problem. The bell-shaped primitive is a velocity profile, and scalar speed does not depend on the choice of reference frame; onset and duration can therefore be recovered from the speed trace alone, where only the temporal factor of $\rho_{i,j}$ limits what can be resolved. The spatial scale is different. It cannot be read from speed, which discards direction, and must instead be recovered from the full trajectory. A method that fits only in velocity must integrate to return to position, and the integration accumulates small velocity errors into a position drift that grows with the duration of the movement --- the behaviour of the Gowda baseline on the long-horizon tasks (Table~\ref{tab:combined_metrics_all}). For robot learning from demonstration, where the reconstructed trajectory is the product of interest, such drift is disqualifying. Sub-ID therefore solves for amplitude in the position domain as its final step. This sacrifices the coordinate invariance of the earlier stages; the spatial error, however, remains bounded even over long movements.

The choice of domain also bears on the noise model. Solving for amplitude by least squares in position is the maximum-likelihood estimate only when the noise on position is white. Human movement does not meet this condition. If the noise corrupting velocity commands is white, the noise it produces in position is Brownian: a random walk whose variance grows linearly with time~\cite{Tessari_2024}. The drift seen in velocity-only methods is therefore not only a numerical artifact of integration; it is in part this same Brownian process, which is why it grows with the duration of the movement. Sub-ID's position-domain fit removes the drift, but in doing so it adopts an objective that is not the maximum-likelihood estimate under Brownian noise. The maximum-likelihood objective would weight the position residual by the covariance of the random walk, which, in the limit, returns the problem to the velocity domain. In the future, a model that treats measurement noise and motor noise together would close this gap and is a natural next step for decompositions intended to recover information closer to the motor command itself.

\subsection{Conclusion}
Submovement decomposition has described human movement for over a century, but without a way to know when a given decomposition can be trusted. Sub-ID supplies that missing piece: a spatiotemporal identifiability criterion, derived from the Cram\'er--Rao bound, that states when submovement amplitudes are recoverable and surfaces its own failure when they are not. Built on this criterion, the method recovers ground-truth parameter distributions where exact search is intractable, and scales to real three-dimensional, long-horizon movements that existing methods cannot decompose without drifting or over-segmenting. In doing so, it turns submovement decomposition from a descriptive fit into a measurement with a known operating range --- and a step toward reading how the nervous system organises movement, rather than only reproducing it.

\section{Methods}

We modelled planar and 3D spatial trajectories, $P(t) \in \mathbb{R}^{D}$ where $D$ is either 2 or 3, as the temporal integration of a resulting tangential velocity, composed of $K$ scaled basis functions:
\begin{equation}
P(t) = P_0 + \sum_{k=1}^{K} \mathbf{w}_{k} \int_{t_{\text{init}}}^{t} \phi(s; \theta_k) ds,
\end{equation}
where $P_0$ represented the initial position, $\mathbf{w}_{k} \in \mathbb{R}^{d}$ was the spatial scale vector that defined the displacement of the $k$-th submovement ($\mathbf{w}_{k} \phi(t; \theta_k)$), and $\phi(t; \theta_k)$ was its normalized basis function, parameterized by the vector $\theta_k$ (encapsulating onset time $t_{0,k}$, duration $T_k$, and shape parameters). Our algorithm aimed to define the parameter vectors $\theta_k$ and scaling vectors $\mathbf{w}_{k}$ whose optimal combination enabled the efficient decomposition of trajectories in both the velocity and spatial domains.

\begin{figure}[ht]
    \centering
    \includegraphics[width=1.0\linewidth]{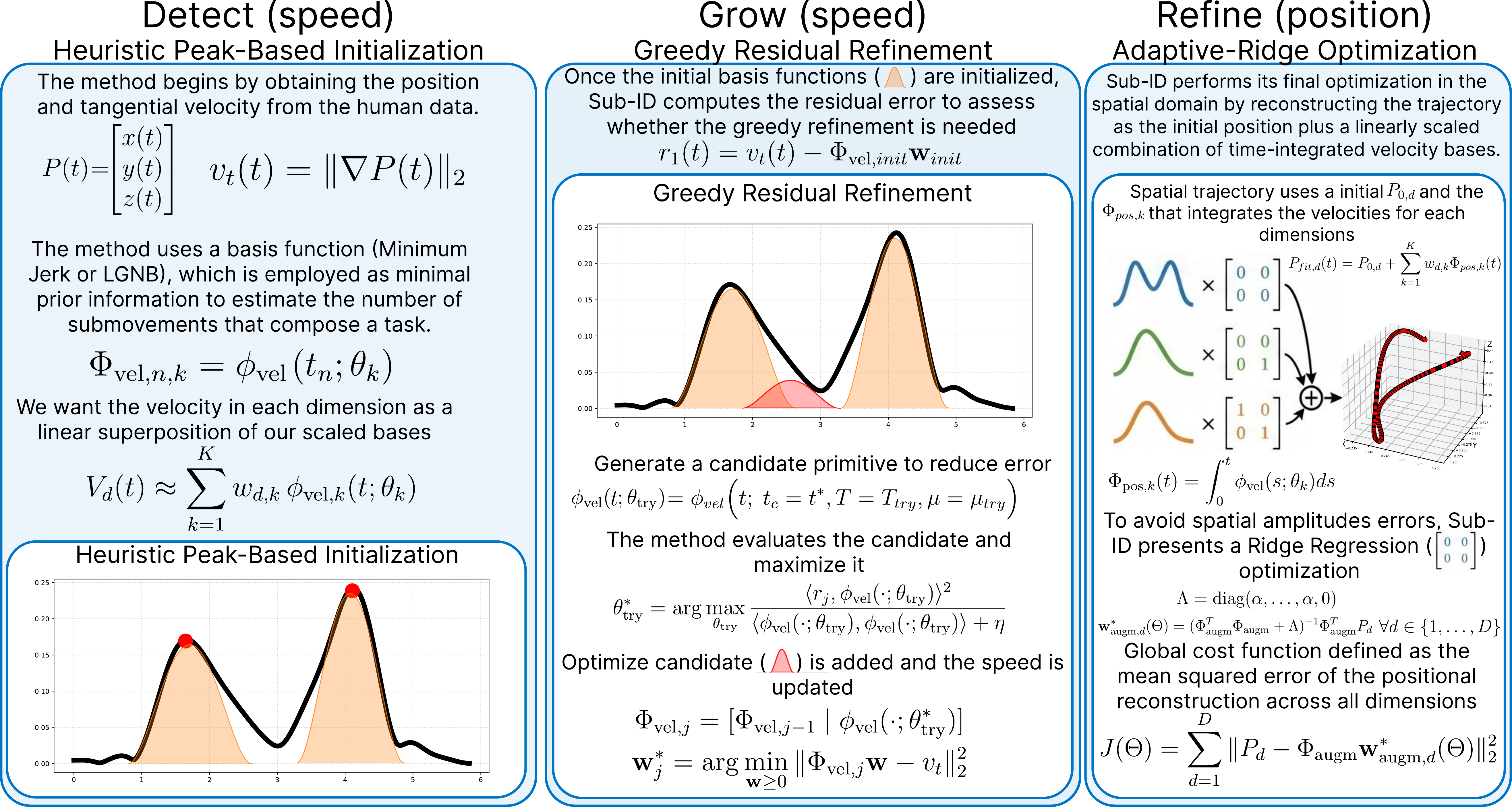}
    \caption{Algorithmic pipeline of the Sub-ID framework. The method begins with \textbf{Heuristic Peak-Based Detection}, extracting the scalar tangential velocity from the human trajectory to identify the initial set of basis functions. The system then evaluates the residual error to trigger \textbf{Greedy Residual Refinement}, where new candidate primitives ($\phi_{\text{vel}}(t;\theta_{\text{try}})$) are iteratively generated, evaluated, and injected to minimize the remaining velocity error. Finally, the \textbf{Adaptative-Ridge Optimization} stage translates these temporal bases into the spatial domain via numerical integration ($\Phi_{\text{pos},k}(t)$). A Ridge Regression approach is applied to robustly compute the spatial amplitudes across all dimensions, ensuring numerical stability, preventing spatial amplitude errors, and finalizing the optimal spatio-temporal reconstruction.}
    \label{fig:GeneralScheme}
\end{figure}

We developed a hierarchical decomposition method that balanced heuristic initialization with a global optimization, linking coordinate-invariant speed with the spatial domain $P(t)$ (Fig.~\ref{fig:GeneralScheme}). The pipeline identified temporal boundaries $(t_{0,k}, t_{1,k})$ and spatial scaling vectors $\mathbf{w}_{k}$ through three stages. The first, \textbf{Detect (speed) Heuristic Peak-Based Initialization}, identified dominant speed transients through peak detection, providing physically plausible initial estimates for each submovement. The second, \textbf{Grow (speed) Greedy Residual Refinement}, iteratively added basis functions to minimize the residual between the reconstructed and observed speed profiles, capturing subtle features without overfitting. The third, \textbf{Refine (position) Adaptive-Ridge Optimization}, simultaneously refined all submovements in the spatial domain, where $L_2$ regularization prevented redundant scaling vectors.

\subsection{Basis functions description}\label{BasisFunctionsDescription}

Numerous basis function choices exist for decomposing human movements, and selecting an appropriate basis function was the subject of substantial prior work. For example, support-bounded Log-normal functions were found to provide the best fit among 23 candidate basis functions \cite{plamondon1993}. However, selecting or comparing basis-function structures was beyond the scope of this work. Instead, our method was designed to be agnostic to the choice of basis function and could therefore be applied to different movement representations.

To demonstrate this flexibility, we evaluated our method using two established basis-function families that differed in their number of degrees of freedom. First, we used the Minimum Jerk basis functions \cite{hogan1984}, which were shown to describe a broad class of voluntary movements using only three parameters. Second, we used the support-bounded Log-normal basis functions~\cite{plamondon1993}, which provided greater descriptive flexibility with five parameters.

\emph{Minimum Jerk:} Based on the principle of maximum smoothness~\cite{hogan1984,flash1985}, this profile minimized the third derivative of position. The velocity profile for a basis function with duration $T_k = t_{1,k} - t_{0,k}$ was defined as:
\begin{equation}\label{Eq: MinJerk}
\phi(t; \theta_k) = \frac{1}{T_k} \left( 30\tau^2 - 60\tau^3 + 30\tau^4 \right), \quad \text{where} \quad \tau = \frac{t - t_{0,k}}{T_k} \in [0, 1].
\end{equation}
MinJerk was parameterized by onset time and duration only ($\theta_k = \{t_{0,k}, T_k\}$), making the grid search one-dimensional.

\emph{Bounded Log-Normal:} Inspired by the Kinematic Theory of Rapid Human Movements~\cite{plamondon1995kinematic,plamondon1995kinematicII}, this model captured the inherent asymmetry of human motion. While standard log-normal distributions possessed asymptotic tails extending to infinity, we utilized the \emph{bounded-support lognormal} (LGNB) model~\cite{plamondon1993} to prevent positional drift caused by the integration of overlapping infinite tails. Unlike the minimum-jerk model, which was originally defined at the position level, LGNB basis functions are defined directly as bounded velocity pulses over a finite temporal support $[t_{0,k}, t_{1,k}]$:
\begin{equation}\label{Eq:LGNB}
\phi(t; \theta_k) =
\begin{cases}
\dfrac{t_{1,k} - t_{0,k}}
{\sigma_k \sqrt{2\pi} (t - t_{0,k})(t_{1,k} - t)}
\exp\left(
-\dfrac{1}{2\sigma_k^2}
\left[
\ln\left( \dfrac{t - t_{0,k}}{t_{1,k} - t} \right) - \mu_k
\right]^2
\right),
& t_{0,k} \leq t \leq t_{1,k}, \\
0, & \text{otherwise},
\end{cases}
\end{equation}
where $t_{0,k}$ and $t_{1,k}$ define the finite temporal boundaries of the basis function, $\mu_k$ controls the skewness (asymmetry), and $\sigma_k$ governs the kurtosis (width), forming the parameter vector $\theta_k = \{t_{0,k}, T_k, \mu_k, \sigma_k\}$ (where $T_k = t_{1,k} - t_{0,k}$).

\subsection{Step-by-Step Submovement Decomposition Algorithm}\label{StepbyStep}
The process of extracting and modeling motor primitives was divided into three main stages: heuristic peak-based initialization, greedy residual refinement, and adaptive-ridge optimization (Fig.~\ref{fig:GeneralScheme}). Prior to this, we present the signal preprocessing. Full code is available in Accession Codes in Sect.\hyperref[AdditionalInfo]{Additional information}. 

Raw position sequences $P_{\text{raw}}(t) \in \mathbb{R}^D$ were resampled to $100$~Hz via cubic \textit{splines} interpolation and filtered with a 4th-order zero-phase Butterworth low-pass filter. The cut-off frequency was set to $5.0$~Hz, which aligned with standard biomechanical practices for upper-limb manipulation tasks (between $4.0$~Hz to $8.0$~Hz), ensuring the retention of valid physiological micro-adjustments while rejecting instrumental artifacts \cite{winter2009biomechanics}.

To mitigate the risk of local minima in high-dimensional optimization, the algorithm employed a hybrid initialization that decoupled temporal and spatial parameters. We defined a coordinate-invariant scalar profile $v_t[n]$, the scalar target profile, representing the tangential velocity magnitude (speed) at time step $n$:
\begin{equation}
v_t[n] = \Big\| \nabla \mathbf{P}[n] \Big\|_2 = \sqrt{\sum_{d=1}^{D} \left( \nabla P_d[n] \right)^2}
\end{equation}
where $\mathbf{P}[n] = [P_1[n], \dots, P_D[n]]^\top \in \mathbb{R}^D$ denoted the spatial position at time step $n$, and $\nabla$ represented the temporal gradient operator along each spatial dimension $d$. This coordinate-invariant quantity decoupled temporal detection from spatial direction, enabling Heuristic Peak-Based Detection and Greedy Residual Refinement to identify primitive onsets and durations independently of workspace geometry or choice of reference frame.

Let $N$ denote the total number of discrete time steps in the sampled trajectory, and $K$ be the number of extracted primitives. Temporal parameters extracted from the speed profile defined an unscaled velocity basis matrix $\Phi_{\text{vel}} \in \mathbb{R}^{N \times K}$ where $\Phi_{\text{vel}, n,k} = \phi_{\text{vel}}(t_n; \theta_k) \quad \text{for } n=1,\dots,N, \quad k=1,\dots,K$. Velocity in each dimension $d$ was modelled as a linear superposition:
\begin{equation}
V_d(t) \approx \sum_{k=1}^K w_{d,k}\,\phi_{\text{vel}}(t; \theta_k)
\end{equation}
where $V_d(t)$ denoted the continuous velocity component in spatial dimension $d$, $w_{d,k}$ represented the dimensional amplitude, and the approximation symbol reflected the inherent residual error when modelling empirical data with a finite set of $K$ primitives. To prevent steady-state error in position resulting from Brownian processes in position~\cite{Tessari_2024}, our framework optimized directly in the position domain. The reconstructed spatial trajectory $\mathbf{P}_{\text{fit}}$ was formulated using an initial position intercept $P_0 \in \mathbb{R}^D$:
\begin{equation}
\mathbf{P}_{\text{fit}} = \mathbf{1_{N}} P_0^\top + \Phi_{\text{pos}}\,\mathbf{W}^\top
\end{equation}
where $\mathbf{1}_N \in \mathbb{R}^N$ is a column vector of ones, $\Phi_{\text{pos},k}(t) = \int_{0}^t \phi_{\text{vel}}(s; \theta_k)\,ds$ integrated the velocity and $\mathbf{W} \in \mathbb{R}^{D\times K}$ contained the scaling coefficients for all dimensions.


Consequently, \textbf{Heuristic Peak-Based Initialization and Greedy Residual Refinement} identified the initial shape parameters $\theta_k$ (onsets, durations) via speed approximation, while the \textbf{Adaptive-Ridge Optimization} refined $\theta_k$ and $\mathbf{W}$ by minimizing the spatial error of $\mathbf{P}_{\text{fit}}$.

This transition to the position domain was necessary because spatial scale vectors $\mathbf{w}_k$ could not be recovered from speed alone, as scalar speed discarded directional information. However, this introduced coordinate dependence: while Heuristic Peak-Based Detection and Greedy Residual Refinement were coordinate-invariant, the position-domain objective $J(\Theta)$ (see Eq.~\ref{J_definition}) was formulated in Cartesian coordinates. Consequently, end-to-end invariance was not preserved; applications in joint-angle space or object-centered frames would require reformulating the global optimization objective for the appropriate coordinate system.

\noindent\textbf{Detect (speed): Heuristic Peak-Based Initialization.}

The simultaneous global optimization of overlapping basis functions over the set of temporal parameters $\Theta = \{\theta_1, \dots, \theta_K\}$, where each $\theta_k$ defines the temporal support of the $k$-th submovement, is a nonconvex, nonlinear problem. To provide a robust initial estimate ($\Theta_{\text{init}}$) and avoid local minima, we proposed a heuristic progressive detection method on the scalar speed profile $v_t(t)$. This \textit{Coarse Pass} isolated dominant submovements through the following stages:

\textit{1) Identification of Local Extrema:} Local maxima in $v_t(t)$ were detected. To filter noise, only peaks exceeding a prominence threshold of $5\%$ of the maximum tangential velocity were retained $\mathbf{M} = \{ k \mid v_t(t_k) \text{ is a peak and } v_t(t_k) \ge 0.05 \cdot \max(v_t) \}$. A minimum temporal distance was also imposed to prevent artifact-driven overlaps.
 
\textit{2) Estimation of Temporal Support:} For each peak $k \in \mathbf{M}$, the width at half maximum was calculated using the left ($t_{\text{left}, k}$) and right ($t_{\text{right}, k}$) crossing times. The center $t_{c, k}$ and empirical width $H_k$ were defined as:
\begin{equation}
H_k = t_{\text{right}, k} - t_{\text{left}, k}, \quad t_{c, k} = \frac{t_{\text{left}, k} + t_{\text{right}, k}}{2}
\end{equation}

\textit{3) Heuristic Expansion and Biological Constraints:} Total duration $T_k$ was estimated using a geometric expansion factor $\gamma \in [1.6, 1.8]$. To ensure neurophysiological plausibility, durations were clipped between $T_{\min}$ ($0.2$~s) and $T_{\max}$ ($5.0$~s):
\begin{equation}
T_k = \max(T_{\min}, \min(T_{\max}, \gamma \cdot H_k))
\end{equation}
The temporal boundaries were defined as $t_{0,k} = t_{c, k} - T_k/2$ and $t_{1,k} = t_{c, k} + T_k/2$.
 
\textit{4) Resolution of Amplitudes and Residual:} Using the temporal parameters $\Theta_{\text{init}}$, the initial weights $\mathbf{w}_{\text{init}}$ were solved via Non-Negative Least Squares (NNLS):
\begin{equation}
\mathbf{w}_{\text{init}} = \arg\min_{\mathbf{w} \ge 0} \| \Phi_{\text{vel, init}} \mathbf{w} - v_t \|_2^2
\end{equation}
NNLS non-negativity was valid in the speed domain because $v_t \ge 0$ by construction; the spatial direction (including negative per-dimension components) was recovered in the global stage via the unconstrained scale vectors $\mathbf{w}_k$.
The resulting residual $r_1(t)$ was computed for further processing in Greedy Residual Refinement: $r_1(t) = v_t(t) - \Phi_{\text{vel, init}} \mathbf{w}_{\text{init}}$.
Pseudocode for Heuristic Peak Detection was provided in Supplementary Material S1.1. 

\noindent\textbf{Grow (speed): Greedy Residual Refinement}

While Heuristic Peak-Based Detection captured dominant submovements, an iterative \textit{Greedy} process on the error signal extracted lower-amplitude micro-movements masked by superposition. In each iteration $j$, the temporal residual $r_j(t)$ was defined as $r_j(t) = v_t(t) - \Phi_{\text{vel}, j-1} \mathbf{w}^*_{j-1}$. The algorithm minimized this residual through residual refinement following these steps:

\textit{1) Stopping Criterion and Error Localization:} The loop continued while the RMS residual remained above the threshold: $\|r_j\|_{\text{RMS}} > \epsilon_{\text{tol}} \max(v_t)$, where $\epsilon_{\text{tol}} \approx 2\%$. The instant of maximum deviation was identified as $t^* = \arg\max_{t} r_j(t)$. A \textit{proximity filter} bypassed $t^*$ if it was critically close ($< \delta_{\text{prox}}$) to an existing primitive's center to avoid redundancies.

\textit{2) Grid Search Exploration:} A discrete search space was evaluated around $t^*$ for durations $T_{\text{try}} \in [0.2, 5.0]\text{ s}$ and, if using LGNB, asymmetries $\mu_{\text{try}}$. Candidate basis functions $\phi_{\text{vel}}(t; \theta_{\text{try}})$ were centered at $t^*$ with support $[t^* - T_{\text{try}}/2, t^* + T_{\text{try}}/2]$.

\textit{3) Topological and Causal Constraints:} Candidates had to satisfy two constraints. First, an \textit{Anti-Nesting} constraint where the primitive could not be entirely subsumed within the temporal support of an existing one. Second, a \textit{Positivity} constraint, where the basis had to explain missing energy, requiring a positive inner product: $\langle r_j, \phi_{\text{vel}}(\cdot; \theta_{\text{try}}) \rangle > 0$.

\textit{4) Score Maximization:} The algorithm selected the parameter candidate $\theta_{\text{try}}^*$ maximizing the squared normalized orthogonal projection:
\begin{equation}
\theta_{\text{try}}^* = \arg\max_{\theta_{\text{try}}} \frac{\langle r_j, \phi_{\text{vel}}(\cdot; \theta_{\text{try}}) \rangle^2}{\langle \phi_{\text{vel}}(\cdot; \theta_{\text{try}}), \phi_{\text{vel}}(\cdot; \theta_{\text{try}}) \rangle + \eta},
\end{equation}
where $\eta = 10^{-9}$ prevented numerical divergence.

\textit{5) Global Update:} The optimal basis evaluated at $\theta_{\text{try}}^*$ was appended to the matrix: $\Phi_{\text{vel}, j} = [\Phi_{\text{vel}, j-1} \mid \phi_{\text{vel}}(\cdot; \theta_{\text{try}}^*)]$. All coefficients were recalculated via NNLS, allowing existing bases to adjust to the new entry:
\begin{equation}
\mathbf{w}^*_j = \arg\min_{\mathbf{w} \ge 0} \| \Phi_{\text{vel}, j} \mathbf{w} - v_t \|_2^2.
\end{equation}
The residual $r_{j+1}(t)$ was recomputed, and the cycle repeated until convergence. Pseudocode for Greedy Residual Refinement was provided in Supplementary Material S1.2.

Fig.~\ref{fig:Fig67}a presents a visual description of Heuristic Peak-Based Initialization and Greedy Residual Refinement. In Heuristic Peak-Based Detection (left image), we observed how the algorithm was able to detect the dominant peak. Using only these peaks, the method was unable to correctly reconstruct the signal. Therefore, a Greedy Residual Refinement was performed (central images), where the algorithm progressively added basis functions and optimized the solution until the error was reduced within the required range.

\noindent\textbf{Refine (position): Adaptive-Ridge Optimization}

Heuristic Peak-Based Initialization and Greedy Residual Refinement provided a precise temporal scaffolding by exploiting the coordinate-invariant nature of tangential velocity, allowing motor command detection regardless of spatial direction. Modeling exclusively in the velocity domain lacked directional information and was vulnerable to \textit{integration drift}, where small fitting errors accumulated into significant spatial deviations.

To address this, our method projected the problem back to the $D$-dimensional spatial domain to optimize the complete set of temporal parameters $\Theta$ by minimizing the reconstruction error of the trajectory $P(t)$. Given the velocity basis matrix $\Phi_{\text{vel}}(\Theta)$, the position-domain matrix $\Phi_{\text{pos}}(\Theta)$ was obtained via integration:
\begin{equation}
\Phi_{\text{pos}, k}(t) = \int_{0}^{t} \phi_{\text{vel}}(s; \theta_k) ds
\end{equation}
To account for the initial spatial offset, an augmented design matrix $\Phi_{\text{augm}} = [\Phi_{\text{pos}}, \mathbf{1}_N] \in \mathbb{R}^{N \times (K+1)}$ was formed by concatenating a column vector of ones. The position refinement followed two subprocesses:

\textit{1) Ridge Regression:} The spatial optimization was solved independently for each dimension $d$ by defining an augmented weight vector $\mathbf{w}_{\text{augm}, d} = [w_{d,1}, w_{d,2}, \dots, w_{d,K}, P_{0,d}]^\top \in \mathbb{R}^{K+1}$, which mapped the dimensional scale factors from $\mathbf{W}$ alongside the initial position intercept. Overlapping basis functions were not strictly orthogonal, so the matrix $\Phi_{\text{augm}}$ could present multicollinearity. If Ordinary Least Squares (OLS) were applied, the optimizer would assign large coefficients of opposite signs to adjacent bases to fit minuscule variations in the residual, producing temporally overlapping primitives that nearly canceled one another. To prevent this, we determined the scales using Ridge Regression~\cite{hoerl1970ridge} ($L_2$ penalty). The solution per dimension $d$ was:
\begin{equation}
\mathbf{w}_{\text{augm}, d}^*(\Theta) = (\Phi_{\text{augm}}^T \Phi_{\text{augm}} + \Lambda)^{-1} \Phi_{\text{augm}}^T P_d
\end{equation}
where $\Lambda \in \mathbb{R}^{(K+1) \times (K+1)}$ was the diagonal regularization matrix $\Lambda = \text{diag}(\alpha, \dots, \alpha, 0)$. The last element was $0$ so that the initial position $P_0$ was not penalized.

To strike a balance between identifiability and reconstruction fidelity, our framework dynamically computed the minimum necessary penalty directly from the system's condition number. While the spatial scales were solved in the position domain, temporal aliasing was a velocity-domain phenomenon. We therefore evaluated the system's conditioning using the normalized velocity basis matrix $\Phi_{\text{norm}}$, where each column was scaled by its $L_2$ norm. Let $\mathbf{G} = \Phi_{\text{norm}}^T \Phi_{\text{norm}}$ be the normalized Gram matrix; we computed its eigenvalues to determine the current condition number $\kappa(\mathbf{G}) = \lambda_{max}(\mathbf{G})/\lambda_{min}(\mathbf{G})$.

Applying Tikhonov regularization~\cite{tikhonov1977solutions} to this Gram matrix yielded a closed-form expression for the minimum ridge penalty $\lambda_{min}^*$ that guaranteed $\kappa(\mathbf{G} + \lambda \mathbf{I}) \le \kappa_{max}$, where $\kappa_{max}$ was the maximum tolerable condition number (empirically set to $100$):
\begin{equation}
\lambda_{min}^* = \sigma^2 \cdot \frac{\lambda_{max}(\mathbf{G}) - \kappa_{max}\lambda_{min}(\mathbf{G})}{\kappa_{max} - 1},
\label{eq:lambda_min}
\end{equation}
where $\sigma^2$ was the average squared $L_2$ norm of the unnormalized velocity kernels. This closed form followed directly from requiring $\kappa(\mathbf{G} + \lambda \mathbf{I}) \le \kappa_{max}$ and solving for $\lambda$ in terms of the eigenvalues of $\mathbf{G}$~\cite{tikhonov1977solutions}.

The spatial ridge parameter $\alpha$ was then set directly to this analytical value, adaptively activating the penalty only when multicollinearity threatened numerical stability:
\begin{equation}
\alpha = 
\begin{cases} 
\lambda_{\min}^*, & \text{if } \kappa(\mathbf{G}) > \kappa_{\max} \\ 
0, & \text{if } \kappa(\mathbf{G}) \le \kappa_{\max}
\end{cases}
\end{equation}
When $\kappa(\mathbf{G}) \le \kappa_{\max}$, the basis functions were deemed sufficiently independent, ensuring that no artificial spatial bias was introduced ($\alpha = 0$). Conversely, when ill-conditioning occurred, $\alpha = \lambda_{\min}^*$ dynamically guaranteed that the system remained bounded by $\kappa_{\max}$.


Because $\mathbf{G}$ was available analytically from the current temporal parameters $\Theta$ at each iteration, this adaptive $\alpha$ was computed dynamically without computational overhead. This eliminated the need for heuristic coalescence or fusion algorithms (\textit{Base Merging}) applied a posteriori. By penalizing extreme magnitudes only when temporal aliasing required it, the optimization prevented adjacent primitives from converging toward the same temporal region with opposite signs. In over-parameterization scenarios, Ridge regression deactivated redundant submovements by shrinking their amplitudes to near zero, so the \textit{L-BFGS-B} optimization converged to a sparse, well-conditioned decomposition.

\textit{2) Global Objective Function:} Once the optimal $\Theta$-dependent scales were guaranteed by the Ridge filter, the global cost function was defined as the mean squared error of the positional reconstruction across all joint spatial dimensions:
\begin{equation}\label{J_definition}
J(\Theta) = \sum_{d=1}^{D} \| P_d - \Phi_{\text{augm}} \mathbf{w}_{\text{augm}, d}^*(\Theta) \|_2^2
\end{equation}
We minimized $J(\Theta)$ iteratively using the \textit{L-BFGS-B} quasi-Newton gradient optimization algorithm, subject to boundary constraints on the temporal parameters (ensuring causality and biological duration limits).

Fig.~\ref{fig:Fig67}b illustrates the global optimization process using MinJerk trajectories, demonstrating that while both Ridge Regression and Ordinary Least Squares accurately reconstructed position and velocity, Ridge Regression was essential for mitigating the Basis Cancellation Phenomenon. As shown in the $\dot x_2$ and $\dot x_3$ comparisons, the $L_2$ penalty ensured parameter identifiability by preventing the generation of simultaneous, opposing basis functions and the redundant nesting of primitives.

\begin{figure}[ht]
    \centering
    \includegraphics[width=1.0\linewidth]{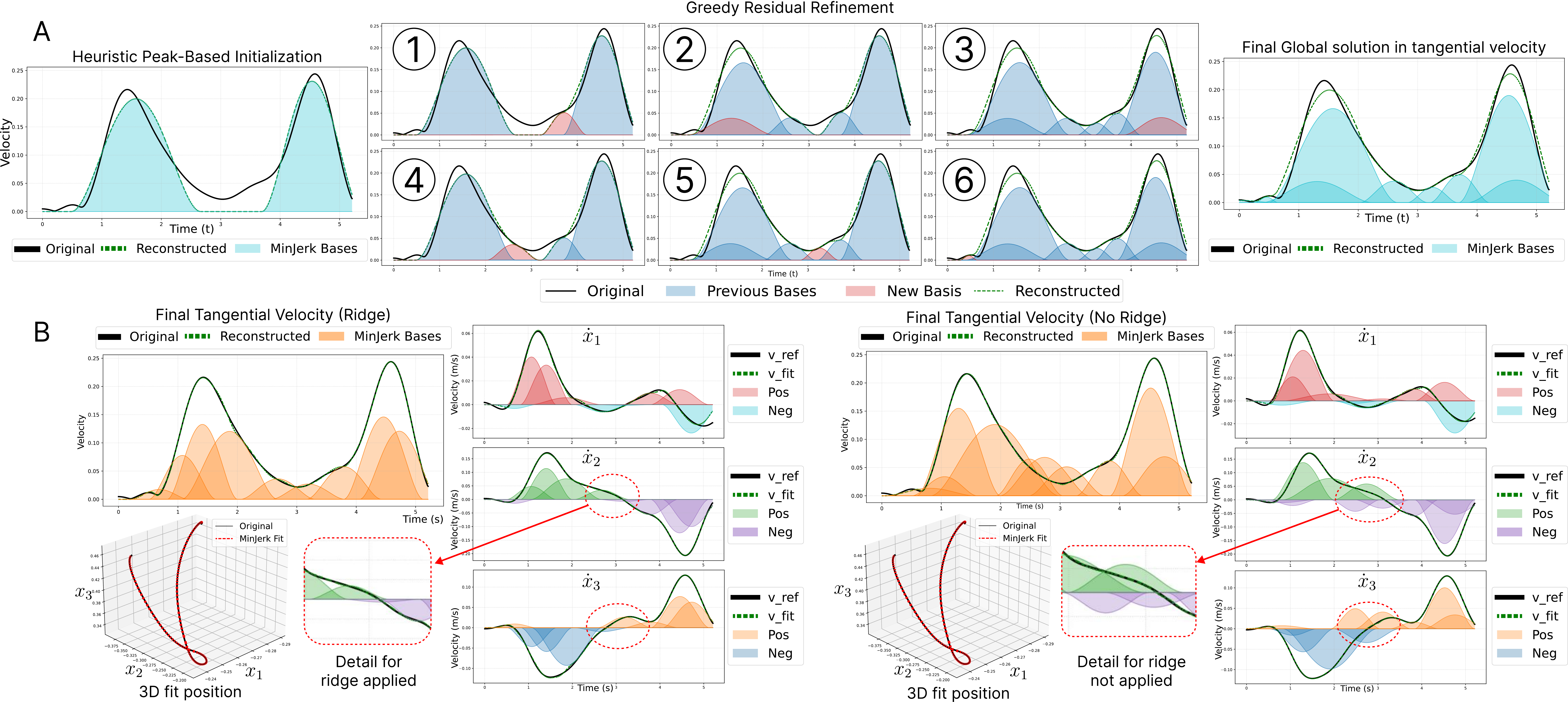}
    \caption{\textbf{(A)}. \textbf{Temporal initialization and greedy residual refinement.} (Left) \emph{Heuristic Peak-Based Initialization} algorithm successfully identifies the most dominant kinematic submovements (light blue), providing a coarse initial fit. Because these primary peaks are insufficient to fully capture the complex movement, a significant residual error remains. (Center) The \emph{greedy refinement process} iteratively locates the maximum residual error and progressively introduces new basis functions (red) to explain the missing kinematic intensity. In each iteration, the amplitudes of all previously discovered bases (blue) are simultaneously re-optimized. (Right) The \emph{final converged solution} accurately reconstructs the scalar tangential velocity profile. This refined temporal scaffolding serves as the robust starting point for the subsequent global optimization in the position domain. \textbf{(B)} \textbf{Impact of Ridge Regression on basis cancellation.} Comparison of the spatio-temporal global optimization of a 3D trajectory with $L_2$ regularization (Left) and without (Right). Both approaches reconstruct the 3D spatial position and tangential velocity, but the unregularized solution exhibits severe multicollinearity: multiple nested primitives emerge simultaneously with extreme opposite signs (visible in $\dot x_2$ and $\dot x_3$, marked in dashed red lines), nearly cancelling one another to fit small residual variations. Ridge Regression (Left) penalizes these extreme amplitudes, yielding a stable, sparse decomposition without altering the total number of basis functions.}
    \label{fig:Fig67}
\end{figure}

\subsection{Identifiability and Spatiotemporal Kernel Correlation}
\label{sec:identifiability}

The identifiability of a submovement decomposition was governed by how distinguishable the basis functions were from one another. We measured this with pairwise kernel correlations.

For two submovements $i$ and $j$ characterized by parameter sets $\theta_i$ and $\theta_j$ (comprising onset times, durations, and asymmetries), their temporal basis functions were denoted as $\phi_{\text{vel}, i}(t) = \phi_{\text{vel}}(t; \theta_i)$ and $\phi_{\text{vel}, j}(t) = \phi_{\text{vel}}(t; \theta_j)$. The temporal kernel correlation between primitives $i$ and $j$ was defined as the normalized inner product of their corresponding velocity basis functions:
\begin{equation}
\rho_{i,j}^{\text{temp}} = \frac{\langle\, \phi_{\text{vel}, i},\, \phi_{\text{vel}, j}\, \rangle}{\|\phi_{\text{vel}, i}\|_2 \, \|\phi_{\text{vel}, j}\|_2}
\label{eq:kernel_corr}
\end{equation}
When two basis functions coincided ($i = j$), $\rho_{i,i}^{\text{temp}} = 1$; as their temporal support separated or their shape parameters diverged, $|\rho_{i,j}^{\text{temp}}|$ dropped toward zero. In $D$ dimensions, each primitive $i$ had an associated unit-norm spatial direction vector $\mathbf{e}_i \in \mathbb{R}^D$, and the total spatiotemporal kernel correlation between submovements $i$ and $j$ was given by:
\begin{equation}
\rho_{i,j} = \cos(\psi_{i,j}) \cdot \rho_{i,j}^{\text{temp}} = (\mathbf{e}_i^\top \mathbf{e}_j) \cdot \frac{\langle\, \phi_{\text{vel}, i},\, \phi_{\text{vel}, j}\, \rangle}{\|\phi_{\text{vel}, i}\|_2 \, \|\phi_{\text{vel}, j}\|_2}
\label{eq:spatiotemporal_corr}
\end{equation}
where $\psi_{i,j}$ was the angle between spatial direction vectors $\mathbf{e}_i$ and $\mathbf{e}_j$. Spatial orthogonality ($\cos\psi_{i,j} = 0$) yielded $\rho_{i,j} = 0$ regardless of temporal overlap, whereas spatial alignment ($\cos\psi_{i,j} = 1$) recovered the purely temporal kernel correlation. In practice, for $D$-dimensional trajectories, temporal parameters were initialized on the scalar speed profile $v_t(t) = \|\mathbf{v}(t)\|_2$. Spatial direction vectors $\mathbf{e}_i$ and submovement magnitudes $w_i$ were subsequently recovered by solving independent 1D Ridge regressions along each spatial coordinate axis and normalizing the resulting weight vector $\mathbf{w}_i = [w_{1,i}, \dots, w_{D,i}]^\top$. A detailed explanation is provided in Supplementary Section S5.4.

The collection of all pairwise correlations formed the normalized Gram matrix $\mathbf{G} \in \mathbb{R}^{K \times K}$ with entries $\mathbf{G}_{ij} = \rho_{i,j}$. Numerically, as implemented in Sub-ID, given the velocity basis matrix $\Phi_{\text{vel}} = [\phi_{\text{vel}, 1}, \dots, \phi_{\text{vel}, K}] \in \mathbb{R}^{N \times K}$, each column was normalized by its $L_2$ norm to construct $\Phi_{\text{norm}}$, whereupon the Gram matrix was computed directly as $\mathbf{G} = \Phi_{\text{norm}}^\top \Phi_{\text{norm}}$. Its condition number $\kappa(\mathbf{G}) = \lambda_{\max}(\mathbf{G})/\lambda_{\min}(\mathbf{G})$ quantified how close the system of basis functions was to linear dependence: $\kappa(\mathbf{G}) = 1$ when all primitives were mutually orthogonal, and $\kappa(\mathbf{G}) \to \infty$ when any pair became collinear ($\rho_{i,j} \to 1$).

This conditioning had a direct statistical consequence for amplitude recovery. Under the linear-Gaussian model used by Sub-ID, assuming fixed temporal parameters $\Theta$ subject to additive noise $\boldsymbol{\varepsilon} \sim \mathcal{N}(\mathbf{0}, \sigma_n^2 \mathbf{I})$, the Cram\'er--Rao bound yielded:
\begin{equation}
\mathrm{Var}(\hat{\mathbf{w}}) \ge \sigma_n^2 \, \mathbf{D}_{\text{norm}}^{-1} \mathbf{G}^{-1} \mathbf{D}_{\text{norm}}^{-1},
\label{eq:crb}
\end{equation}
where $\mathbf{D}_{\text{norm}} = \text{diag}(\|\phi_{\text{vel}, 1}\|_2, \dots, \|\phi_{\text{vel}, K}\|_2)$. Worst-case amplitude variance scaled directly with $\kappa(\mathbf{G})$, meaning that high pairwise correlations $\rho_{i,j}$ directly degraded amplitude recovery accuracy. In the degenerate case $\rho_{i,j} = 1$, the matrix $\mathbf{G}$ became singular and amplitude recovery was information-theoretically impossible \cite{donoho2003optimally}. Equation~\ref{eq:crb} was the standard Cram\'er--Rao bound for this linear-Gaussian model; its variance floor was governed entirely by the conditioning of $\mathbf{G}$, so the same condition number $\kappa(\mathbf{G})$ that flagged ill-posedness set the achievable accuracy of amplitude recovery.

In practice, Sub-ID computed $\mathbf{G} = \Phi_{\text{norm}}^\top \Phi_{\text{norm}}$ numerically at each iteration and evaluated $\kappa(\mathbf{G}) \le \kappa_{\max}$ as the diagnostic for well-conditioning. The adaptive ridge parameter $\alpha$ was determined analytically from $\mathbf{G}$ via Eq.~\eqref{eq:lambda_min}, directly embedding this identifiability criterion into the optimization framework.

We characterized the identifiability of amplitude estimation given fixed temporal parameters; joint identifiability of onset times, durations, and asymmetries was nonconvex and was not addressed formally here, but Sub-ID's behaviour in that regime was demonstrated empirically in Section~\ref{sect:garcefulFailure}.

\subsection{Experimental Methods}
To evaluate Sub-ID's performance and scalability, we conducted experiments on synthetic and real kinematic data.

To evaluate performance against an absolute ground truth, we conducted \textbf{Synthetic Data Experiments} using artificial trajectories with known basis function parameters (onset, duration, amplitude, and asymmetry). Importantly, sequential primitives were generated with high temporal overlap by anchoring the baseline onset of each subsequent submovement around the peak velocity of the preceding one (i.e., half-duration $D/2$ for symmetric minimum-jerk profiles~\cite{flash1985coordination}). A stochastic temporal offset $\Delta t$ was added to this anchor point to introduce variability across submovements, representative of the dense, highly overlapping submovement control observed in motor execution~\cite{milner1992model, doeringer1998serial}. This controlled setting allowed us to benchmark our algorithm against \text{Scattershot}~\cite{friedman2019observation}, a search-space reduction method (Gowda et al.~\cite{gowda2015}), and \text{SSSUMO}~\cite{rudakov2025}, under increasing complexity. By systematically varying the number of overlapping primitives and parameter distributions, we evaluate how primitive overlap affects identifiability and numerical stability across different methods.


We conducted \textbf{Real Data Experiments} to quantify the algorithm's practical applicability and biological plausibility. By applying our framework to datasets ranging from 2D planar reaching to complex 3D manipulation tasks, we quantified the extent to which the method robustly scales to noisy, high-dimensional human kinematics. 


\vspace{3mm}
\noindent\textbf{Synthetic Data Experiments}

To evaluate the algorithm's precision in recovering latent parameters, we designed a synthetic environment that emulates human kinematic complexity. Rather than using random parameters, our generator utilizes realistic statistical distributions for submovement duration ($D$), peak amplitude ($A$), and refractory period ($L$). These distributions are derived from prior kinematic analyses of real 2D and 3D manual tasks across multiple users to establish the nominal parameters of human execution.


\begin{figure}[ht]
    \centering
    \includegraphics[width=1.0\linewidth]{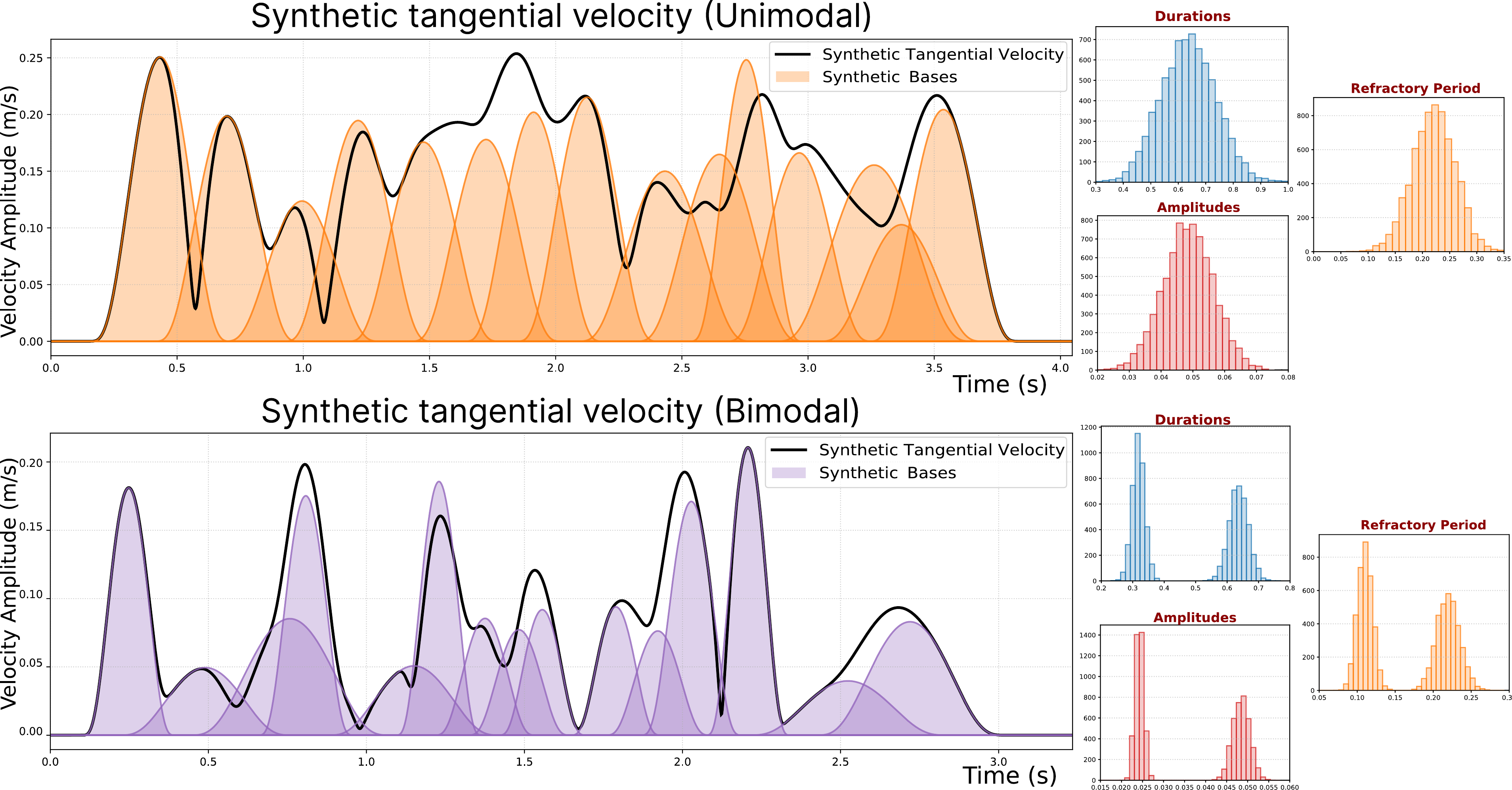}
    \caption{Unimodal and bimodal synthetic data. The central panels depict the temporal organization of 15 overlapping basis functions generated. The right panels show the empirical distributions of duration, amplitude, and refractory period, parameterized from human demonstrations collected in extended 2D and 3D tasks. In the bimodal setting, these distributions exhibit a clear separation into two distinct modes, which is consistently reflected in the structure of the reconstructed tangential velocity profiles and the corresponding basis function activations.}
    \label{fig:UniandBi}
\end{figure}


We evaluated the algorithm across different complexity scenarios. Unlike most classical algorithms, whose computational tractability and accuracy is typically demonstrated only on short tasks with few primitives, Sub-ID assesses the system's robustness against increasing levels of complexity. To this end, three states were defined based on the number of basis functions: Low, medium, and high complexity. The \textit{Low Complexity ($K=3$)} represented simple ballistic movements or short-horizon tasks. This is the typical range within which many previous works remain robust and computationally tractable. The \textit{Medium Complexity ($K=8$)} simulated moderately complex movement sequences. The \textit{High Complexity ($K=15$)} represented comparatively longer-horizon tasks characterized by multiple kinematic corrections and a high degree of temporal overlap. 

Decomposing extended trajectories with over 15 submovements is often computationally prohibitive for state-of-the-art methods. This tiered $K$-level analysis identifies the breaking point of existing proposals, particularly when high primitive density hinders parameter identifiability. To evaluate algorithmic robustness across varied kinematic scales, we implement two distribution modes for each level of $K$: unimodal and bimodal. The \textit{Unimodal Mode} parameters were sampled from normal distributions centered on observed human medians ($\tilde{D}$, $\tilde{A}$, $\tilde{L}$), representing standard coordinated movements. The \textit{Bimodal Mode} Parameters were sampled from a bimodal distribution, combining nominal values with a secondary set of scales modified by 50\%. This structural heterogeneity tested the algorithm's capacity to resolve highly overlapping primitives with significantly different scales and durations without coalescence.

The Pseudocode for the generation of that synthetic data is presented in Supplementary Materials in S1.3. Fig.~\ref{fig:UniandBi} illustrates the time-ordered generation of 15 overlapping basis functions for tangential velocity. The right-hand panels display the distributions of duration, amplitude, and refractory period derived from human 2D/3D task data. In the bimodal case, these distributions—and their corresponding velocity basis functions—clearly bifurcate into two distinct groups.

Additionally, we introduced the \textbf{Extreme Overlapping Mode} (or Temporal Aliasing) to investigate the mathematical limits of the decomposition framework. By minimizing the refractory period such that $L \ll D$, the velocity profile suffered from severe temporal aliasing, masking individual kinematic landmarks. This resulted in a highly ill-posed inverse problem where adjacent basis functions become nearly linearly dependent. Mathematically, it becomed impossible to distinguish between two highly collinear primitives and a single larger one, as both yield identical velocity profiles. This scenario illustrated the theoretical limits of identifiability, where extreme aliasing prevents the unique decoupling of underlying primitives without imposing non-physiological constraints.

\vspace{3mm}
\noindent\textbf{Real Data Experiments}

Following synthetic validation, we evaluated the algorithm's performance on human kinematics to test its robustness against real-world signal distortions, such as sensorimotor noise. The experiments were categorized by dimensionality (2D vs. 3D) and temporal horizon (short vs. long). To benchmark against state-of-the-art methods, which primarily operate on short tasks, we utilized two public databases; for long-horizon tasks, we utilized two proprietary datasets in both 2D and 3D environments.

\textit{1) Short-Horizon Tasks: Existing Datasets.} For this evaluation, short-horizon tasks were defined as rapid movements reconstructible with a limited number of basis functions (typically $K < 8$). We included these scenarios to test whether algorithms can accurately capture fast kinematics without overfitting or generating spurious submovements~\cite{rohrer2003}, as baseline methods were theoretically expected to perform well in this tractable regime.

To evaluate the method on \emph{2D short tasks}, we utilized the \textit{Character Trajectories Data Set} from the UCI Machine Learning Repository, a standard benchmark in motor primitive analysis~\cite{williams2006, williams2007modelling} presented by Ben H. Williams. The dataset contains pen tip trajectories recorded at $200$\,Hz via a WACOM digitizing tablet, retaining spatial dimensions ($x, y$), force, and stroke velocity across approximately $150$ samples per character. Handwriting represents an ideal case study for motor control, as it necessitates the precise temporal localization and sequential concatenation of rapid, highly coordinated ballistic movements.

    
Next, to evaluate the algorithm on \emph{3D short tasks}, we utilized the \textit{Object Moving} task~\cite{grimme2012naturalistic} from the SSSUMO framework validation~\cite{rudakov2025}. This 3D task involved moving a cylinder while avoiding obstacles of variable heights, introducing complex dynamics such as lifting and translating. This dataset was critical because 3D tangential velocity integrates more spatial components than planar tasks, often causing traditional methods to fail or incur high computational costs. The dataset included $1600$ examples recorded at $110$\,Hz using a Visualeyez VZ 4000 system with an infrared LED (IRED) attached to the object. Signals were processed using a table-anchored reference frame and a $5.5$\,Hz low-pass filter~\cite{raket2016separating}.

\textit{2) Long-Horizon Tasks: Custom Datasets.} Short-duration benchmarks were standard in the literature, yet real-world manipulation often involves extended horizons characterized by continuous corrections and complex obstacle avoidance. These long-duration tasks generated multi-peaked, overlapping velocity profiles requiring many primitives ($K > 15$), where traditional algorithms typically fail due to integration drift and spurious decompositions. To test the scalability and robustness of our framework in these continuous scenarios, we collected two custom datasets inspired by the PushT task from the Diffusion Policy benchmark \cite{Chi_2024}.

This tasks required the user to push a T-shaped block from randomized initial configurations into a fixed target zone using the tip of a tool. To capture distinct neuromotor strategies and evaluate dimensional scalability, we recorded two variants a 2D Planar PushT and a 3D Spatial PushT.

In the \textit{2D Planar PushT} task, users were instructed to keep the tip of the tool on the plane of the T-block. Specifically, they were not allowed to lift the tool over the T-block to move from one side to the other; instead, the user had to continuously navigate around the contours of the piece to reposition it. This restriction forces the execution of continuous, high-precision corrective submovements in approximately two-dimensional space, generating a high density of overlapping primitives. Next, in the \textit{3D Spatial PushT} variant, users were allowed to lift the tool, breaking the 2D plane to cross over the T-block and reposition the tool on the other side. This introduces vertical motion ($Z$-axis motion), requiring the algorithm to decouple the primitives across three spatial dimensions.

To ensure kinematic consistency across all recordings, the execution time for each trial was bounded to a maximum duration of $15$ seconds. Each user performed a collection of 50 tasks in both 2D/3D. Furthermore, all executions strictly began from a state of absolute rest (zero initial velocity), with the tool physically resting at the starting point of the task, and were performed within a restricted workspace measuring $0.55$\,m per side, as presented in Supplementary Material S3.

\vspace{3mm}
\noindent\textbf{Statistical Analysis and Evaluation Metrics}

To comprehensively evaluate the performance of the decomposition algorithms, we designed an evaluation two sets of metrics: kinematic reconstruction fidelity and the mathematical identifiability of latent structures.

\textit{1) Kinematic Reconstruction Fidelity:} These metrics evaluated the algorithm's ability to fit the kinematics of the original signal. 

To evaluate spatial accuracy, the \textit{Spatial Position Error (Pos Error)} measured integration drift by calculating the Root Mean Square Error (RMSE) between the real trajectory $P(t_n)$ and the reconstructed trajectory $\hat{P}(t_n)$:
\begin{equation}
\text{RMSE}_{pos} = \sqrt{\frac{1}{N} \sum_{n=1}^{N} \left\| P(t_n) - \hat{P}(t_n) \right\|^2},
\end{equation}
where $N$ represents the total number of temporal samples. A value near zero confirms the absence of cumulative spatial reconstruction errors. Furthermore, to assess the temporal dynamics, the \textit{Velocity Reconstruction (Vel $R^2$ \& Vel RMSE)} metrics quantified the explained variance and absolute error of the scalar velocity profile:
\begin{equation}
R^2_{vel} = 1 - \frac{\sum_{n=1}^N (v(t_n) - \hat{v}(t_n))^2}{\sum_{n=1}^N (v(t_n) - \bar{v})^2}, \quad \text{RMSE}_{vel} = \sqrt{\frac{1}{N} \sum_{n=1}^{N} \left( v(t_n) - \hat{v}(t_n) \right)^2},
\end{equation}
where $v(t_n)$ was the empirical tangential velocity, $\hat{v}(t_n)$ was the modelled velocity, and $\bar{v}$ is the mean real velocity. High $R^2_{vel}$ and low $\text{RMSE}_{vel}$ values indicate a precise mathematical capture of kinematic intensity.

\textit{2) Mathematical Identifiability and Structural Complexity:} These metrics evaluated the temporal density of the recovered decomposition and quantify the risk of multicollinearity. Sect.~\ref{sec:identifiability} presents the underlying theory and definition of~$\rho_{i,j}$.

The \textit{Primitive Rate ($K$ \& Rate $K/s$)} indicated the total number of generated basis functions ($K$) and their temporal density (primitives per second) via $K/T_{\text{total}}$, where $T_{\text{total}}$ was the total duration of the analyzed task; high rates increase the risk of overfitting and extreme overlapping. Next, the \textit{Temporal Overlap Metrics (Mean Overlap, Max Overlap, \% Overlap $> 0.8$)} quantified the temporal superposition between two adjacent primitives as the fraction of their intersection relative to the shortest primitive:
\begin{equation}
O_{i,j} = \frac{\max(0, \min(t_{1,i}, t_{1,j}) - \max(t_{0,i}, t_{0,j}))}{\min(D_i, D_j)},
\end{equation}
where $t_{0,(\cdot)}$ and $t_{1,(\cdot)}$ represented the onset and offset times of each submovement, and $D_{(\cdot)}$ was its total duration. We reported the mean, maximum, and percentage of critical interactions (those with $O_{i,j} > 0.8$), where extreme values precipitate the basis cancellation phenomenon~\cite{rohrer2003,rohrer2006}. Finally, the \textit{Spatiotemporal Kernel Correlation ($\rho_{i,j}$)} measured the mean and maximum of $\rho_{i,j}$ across pairs of recovered primitives (Sect.~\ref{sec:identifiability}), where high values flag ill-conditioning of the recovered decomposition.

Supplementary Material Section S5 provides additional mathematical details alongside a summary of all evaluation metrics.

\subsection{Implementation and Adaptation of Comparative Methods}

To ensure a fair, rigorous, and transparent evaluation, the proposed optimization framework was compared against three state-of-the-art submovement decomposition methods presented in detail in the literature: Scattershot~\cite{rohrer2006}, Gowda et al.~\cite{gowda2015}, and SSSUMO~\cite{rudakov2025}. Baseline methods were implemented using their official open-source repositories. Only specific boundary conditions were applied to these codes to standardize the comparative framework with respect to our algorithm. All methods were executed on the same hardware, using an MSI Katana GF66 laptop equipped with an Intel Core i7 processor (4.7 GHz).

\vspace{3mm}
\noindent\textbf{Scattershot (Jason Friedman implementation)}

The Scattershot algorithm\cite{rohrer2006}, implemented via the codebase provided by Jason Friedman in his official repository (Sect.\hyperref[AdditionalInfo]{Additional information}), relies on a highly iterative search to minimize the reconstruction error. To evaluate this method under practical and computationally feasible conditions, we established a strict stopping criterion: the algorithm terminates successfully if the velocity profile reconstruction error drops below $2\%$.


Since Scattershot's unconstrained search can potentially execute indefinitely on complex trajectories without reaching theoretical convergence, we introduced an upper bound on the maximum number of basis functions allowed. If the $2\%$ error threshold is not reached within this limit, the algorithm is forced to halt and returns the configuration that yielded the minimum error up to that point. This adaptation ensures a reasonable balance between reconstruction fidelity and execution time, avoiding unreasonably long computation time while extracting the solution the algorithm can offer. In addition to these modifications, to maximize the algorithm's performance and ensure a fair comparison, we modified its sampling strategy. While several studies suggest that $10$ independent runs are sufficient to obtain robust results, we empirically increased this parameter to $20$ independent inference runs per trajectory. The configuration with the minimum error among these $20$ runs was selected for the final comparison, deliberately providing the utilized Scattershot implementation with a wider margin of opportunity to find its optimal latent representation.

\vspace{3mm}
\noindent\textbf{Search-Space Reduction (Gowda implementation)}

The Search-Space Reduction algorithm proposed by Gowda et al.~\cite{gowda2015}, relies on a heuristic greedy sampling approach coupled with nonlinear optimization to iteratively reconstruct the kinematic profile. Because the original MATLAB implementation (Sect.\hyperref[AdditionalInfo]{Additional information}) was strictly formulated for two-dimensional planar movements, we translated and generalized the algorithmic core into an N-dimensional Python framework to support the 3D operational space required by our robotic manipulation datasets. 

In our implementation, which is release as part of our code,  (see Sect.\hyperref[AdditionalInfo]{Additional information}), the algorithm segmented the trajectory and iteratively allocated new basis functions by sampling initialization times at the local minima of the inverse residual probability distribution. To compute the optimal primitive parameters, the spatial amplitudes across all dimensions were solved efficiently via regularized linear least squares (Ridge regression), while the temporal variables (onset time and duration) were optimized using Sequential Least Squares Programming (SLSQP). To prevent degenerate solutions and unbounded basis stacking, we enforced the authors' original hard constraints, including boundary conditions for durations and a minimum Inter-Spike Interval (ISI) to guarantee a mathematical separation between primitives. The iterative extraction halts dynamically either when the reconstruction error dropped below the $2\%$ target threshold or when the marginal cost improvement between consecutive iterations becomes negligible.

\vspace{3mm}
\noindent\textbf{SSSUMO}

The SSSUMO algorithm \cite{rudakov2025} is a semi-supervised method (Sect.\hyperref[AdditionalInfo]{Additional information}); thus, its authors have conducted various training processes using diverse datasets. Consequently, when executing this algorithm, we utilized the pre-trained models native to the SSSUMO method. Consistent with the model training, all kinematic data evaluated by SSSUMO were resampled at $60\text{ Hz}$.

Unlike Scattershot, SSSUMO operates via inference through a previously trained generative model; therefore, explicit reconstruction error thresholds (such as the $2\%$ limit) cannot be dynamically imposed during the extraction process.

\bibliography{mainOverleaf}

\section*{Acknowledgements}

A.P and R.B were funded by Advanced Mobile dual-arm manipulator for Elderly People Attendance (AMME) (PID2022-139227OB-I00) by Ministerio de Ciencia e Innovacion of Spain.

\noindent J.H. and S.C. were funded in part by Innosuisse — Swiss Innovation Agency, Innovation Project 120.233 IP-ENG. J.H. was also funded in part by the Swiss National Science Foundation (SNSF) Ambizione grant 233246.

\section*{Author contributions statement}

A.P. and J.H. conceived the study and method. A.P. implemented the software, collected the experimental data, and performed the analysis. A.P. and J.H. drafted the manuscript. S.C. contributed to the conceptual development of the method. S.C. and R.B. critically revised the manuscript. All authors reviewed and approved the final manuscript.

\section*{Additional information}
\phantomsection
\label{AdditionalInfo}

\noindent\textbf{Accession codes};

The Sub-ID algorithm developed and proposed in this paper. Available online: \url{https://github.com/AdrianPrados/Sub-ID}

\textit{Character Trajectories Data Set} from the UCI Machine Learning Repository. Available online: \url{https://archive.ics.uci.edu/dataset/175/character+trajectories} 

The Scattershot algorithm by Jason Friedman,  official repository. Available online: \url{https://github.com/JasonFriedman/submovements}

The Search-Space Reduction method based on Gowda et al. implementation. Available online: \url{https://github.com/sgowda/decompose_submovements}

The SSSUMO method. Available online: \url{https://github.com/dolphin-in-a-coma/sssumo}

\section*{Data Availability}
Additional information, including additional experimental results, pseudocodes, and diagrams, are available in the \textbf{Supplementary Data}. The datasets used throughout the experiments are available on Zenodo. Available online: \href{https://zenodo.org/records/21893094?preview=1&token=eyJhbGciOiJIUzUxMiJ9.eyJpZCI6ImY1ODhlYzJkLWMxNDgtNGNlYy05OWU2LTI4ZjIyNDA4YjgzNCIsImRhdGEiOnt9LCJyYW5kb20iOiJkZWM0ZGZiNzMzMWUzZDNiYzlkM2VlMmIxMzRjMTNhYyJ9.enDaC1SwVGoupHzAObSaO1doWeNvyB9TWkg1gd7FiJjD_nJMn0NlVPXEQtcCJRUsKpP0IpyWNNmpKdjQe5dU9w}{Sub-ID Dataset}

\section*{Competing interests}
The authors declare no competing interests.



\end{document}


\flushbottom
\maketitle

\thispagestyle{empty}

\tableofcontents
\newpage 

\section{Pseudocodes}
\subsection{Detect (speed): Heuristic Peak-Based Initialization}

Algorithm~\ref{HeuristicPseudo} outlines the complete code of the process, describing the coarse initialization pass that identifies dominant speed transients and provides the initial parameter estimate $\Theta_{init}$ and initial scaling weights $w_{init}$. 

\begin{algorithm}[H]
\caption{Heuristic Peak Detection}\label{HeuristicPseudo}
\begin{algorithmic}[1]
\STATE \textbf{Input:} Tangential velocity $v_t(t)$, minimum peak height ratio $h_{th} = 0.05$, scaling factor $\gamma$, residual threshold $\epsilon$
\STATE \textbf{Output:} Initial set of primitive parameters $\Theta_{init}$ and scaling weights $w_{init}$


\STATE Identify local maxima indices $P = \{k \mid v_t(t_k) \text{ is a peak and } v_t(t_k) \ge h_{th} \cdot \max(v_t)\}$
\FOR{each detected peak $k \in P$}
    \STATE Find left and right crossing times $t_{left, k}$ and $t_{right, k}$ at relative height
    \STATE Compute empirical width: $W_k = t_{right, k} - t_{left, k}$
    \STATE Compute center time: $t_{c, k} = \frac{t_{left, k} + t_{right, k}}{2}$
    \STATE Estimate duration with bounds: $D_k = \max(D_{min}, \min(D_{max}, \gamma \cdot W_k))$
    \STATE Set temporal bounds: $t_{0,k} = t_{c,k} - D_k/2$ and $t_{1,k} = t_{c,k} + D_k/2$
    \STATE Add $(t_{0,k}, t_{1,k})$ to parameter set $\Theta_{init}$
\ENDFOR
\STATE Construct preliminary basis matrix $\Phi_{vel, init}$ from $\Theta_{init}$
\STATE Solve weights via NNLS: $w_{init} = \arg\min_{w \ge 0} \| \Phi_{vel, init}w - v_t \|_2^2$
\RETURN  $\Theta_{init}$, $w_{init}$
\end{algorithmic}
\end{algorithm}

\newpage
\subsection{Grow (speed): Greedy Residual Refinement}
Algorithm~\ref{ResidualPseudo} describes the greedy refinement pass that captures lower-amplitude primitives
missed by Peak-Based Initialization.

\begin{algorithm}[H]
\caption{Iterative Residual Refinement}\label{ResidualPseudo}
\begin{algorithmic}[1]
\STATE \textbf{Input:} $v_t(t)$, $\Theta_{init}$, $w_{init}$, tolerance threshold $\epsilon_{\text{tol}}$, search grid $\mathcal{T} \times \mathcal{M}$, max primitives $K_{max}$, $\eta = 10^{-9}$
\STATE \textbf{Output:} Expanded primitive parameter set $\Theta$ and optimized scalar weights $w^*$

\STATE $j \gets 1$, $\Phi_{vel, 0} = \Phi_{vel, init}$, and $w^*_0 = w_{init}$
\STATE Compute initial residual: $r_1(t) = v_t(t) - \Phi_{vel, 0} w^*_0$

\WHILE{$\|r_j(t)\|_{RMS} > \epsilon_{\text{tol}} \cdot \max(v_t)$ \AND $|\Theta| < K_{max}$}
    
    \STATE Find time of maximal residual error: $t^* \gets \arg\max_t r_j(t)$
    
    \STATE \textit{// 1. Proximity Filter}
    \IF{$\min_{k \in \Theta} |t_{c,k} - t^*| < \delta_{prox}$}
        \STATE Set $r_j(t) \gets 0$ in the local neighborhood of $t^*$; \textbf{continue} \textit{// Skip to next maximum}
    \ENDIF

    \STATE Initialize $S_{best} \gets -\infty$ and $\phi^*_{vel, try} \gets \emptyset$
    
    \STATE \textit{// 2. Grid Search over parameter space}
    \FOR{each candidate duration $T_{try} \in \mathcal{T}$ (and $\mu_{try} \in \mathcal{M}$)}
        \STATE Generate candidate primitive $\phi_{vel, try}$ centered at $t^*$
        
        \STATE \textit{// 3. Anti-Nesting Topological Constraint}
        \IF{support of $\phi_{vel, try}$ is fully subsumed within any active primitive in $\Theta$}
            \STATE \textbf{continue}
        \ENDIF
        
        \STATE \textit{// 4. Positivity Constraint}
        \IF{$\langle r_j, \phi_{vel, try} \rangle \le 0$}
            \STATE \textbf{continue}
        \ENDIF
        
        \STATE \textit{// 5. Objective Function Evaluation}
        \STATE Compute projection score: $S \gets (\langle r_j, \phi_{vel, try} \rangle^2)/(\langle \phi_{vel, try}, \phi_{vel, try} \rangle + \eta)$
        
        \IF{$S > S_{best}$}
            \STATE $S_{best} \gets S$, $\quad \phi^*_{vel, try} \gets \phi_{vel, try}$
        \ENDIF
    \ENDFOR
    
    \IF{no valid $\phi^*_{vel, try}$ was found}
        \STATE Set $r_j(t) \gets 0$ in the local neighborhood of $t^*$; \textbf{continue}
    \ELSE
    
    \STATE \textit{// 6. Full Velocity Update}
    \STATE Add $\phi^*_{vel, try}$ to the active set $\Theta$
    \STATE Reconstruct velocity basis matrix $\Phi_{vel, j}$ using updated $\Theta$
    \STATE Recompute all weights via NNLS: $w^*_j = \arg\min_{w \ge 0} \| \Phi_{vel, j} w - v_t \|_2^2$
    \STATE Update residual: $r_{j+1}(t) = v_t(t) - \Phi_{vel, j} w^*_j$; $\quad j \gets j + 1$
    \ENDIF

\ENDWHILE

\STATE \Return $\Theta$, $w^*$
\end{algorithmic}
\end{algorithm}

\subsection{ Human-Grounded Synthetic Dataset Generation}
For the generation of the synthetic data, we developed an algorithm that describes the logic for generating the reference parameters (Ground Truth) used for validation. This pseudocode is detailed in Algorithm~\ref{alg:synthetic_gen}.

\begin{algorithm}[H]
\caption{Human-Grounded Synthetic Dataset Generation}\label{alg:synthetic_gen}
\begin{algorithmic}[1]
\REQUIRE Number of primitives $K$, Distribution mode $M \in \{\text{unimodal, bimodal}\}$
\ENSURE Onset parameters $\mathbf{t}_0$, offset parameters $\mathbf{t}_1$, and spatial scales $\mathbf{w}$

\STATE \textbf{Reference parameters (Human Analysis):} $\tilde{T}$, $\tilde{A}$, $\tilde{L}$

\STATE $t_{0,1} \sim \mathcal{N}(0.1, 0.02)$ \COMMENT{Start of the kinematic sequence}
\STATE $T_1 \sim \text{SampleDuration}(\tilde{T}, M)$
\STATE $t_{1,1} \gets t_{0,1} + T_1$

\FOR{$k = 2$ \TO $K$}
    \STATE $L_k \sim \text{SampleLatency}(\tilde{L}, M)$ \COMMENT{Overlap latency between bases}
    \STATE $t_{0,k} \gets t_{0,k-1} + L_k$
    \STATE $T_k \sim \text{SampleDuration}(\tilde{T}, M)$
    \STATE $t_{1,k} \gets t_{0,k} + T_k$
\ENDFOR

\STATE $\mathbf{A} \sim \text{SampleAmplitudes}(\tilde{A}, M)$ \COMMENT{Polar magnitudes $A_k = \|\mathbf{w}_k\|$}
\FOR{$k = 1$ \TO $K$}
    \STATE $\psi_k \sim \mathcal{U}(0, 2\pi)$ \COMMENT{Projection into D-dimensional spatial domain}
    \STATE $w_{x,k} \gets A_k \cos(\psi_k)$, $w_{y,k} \gets A_k \sin(\psi_k)$
\ENDFOR
\RETURN $\mathbf{t}_0, \mathbf{t}_1, \mathbf{w}_x, \mathbf{w}_y$
\end{algorithmic}
\end{algorithm}

\begin{figure}[ht]
    \centering
    \includegraphics[width=1.0\linewidth]{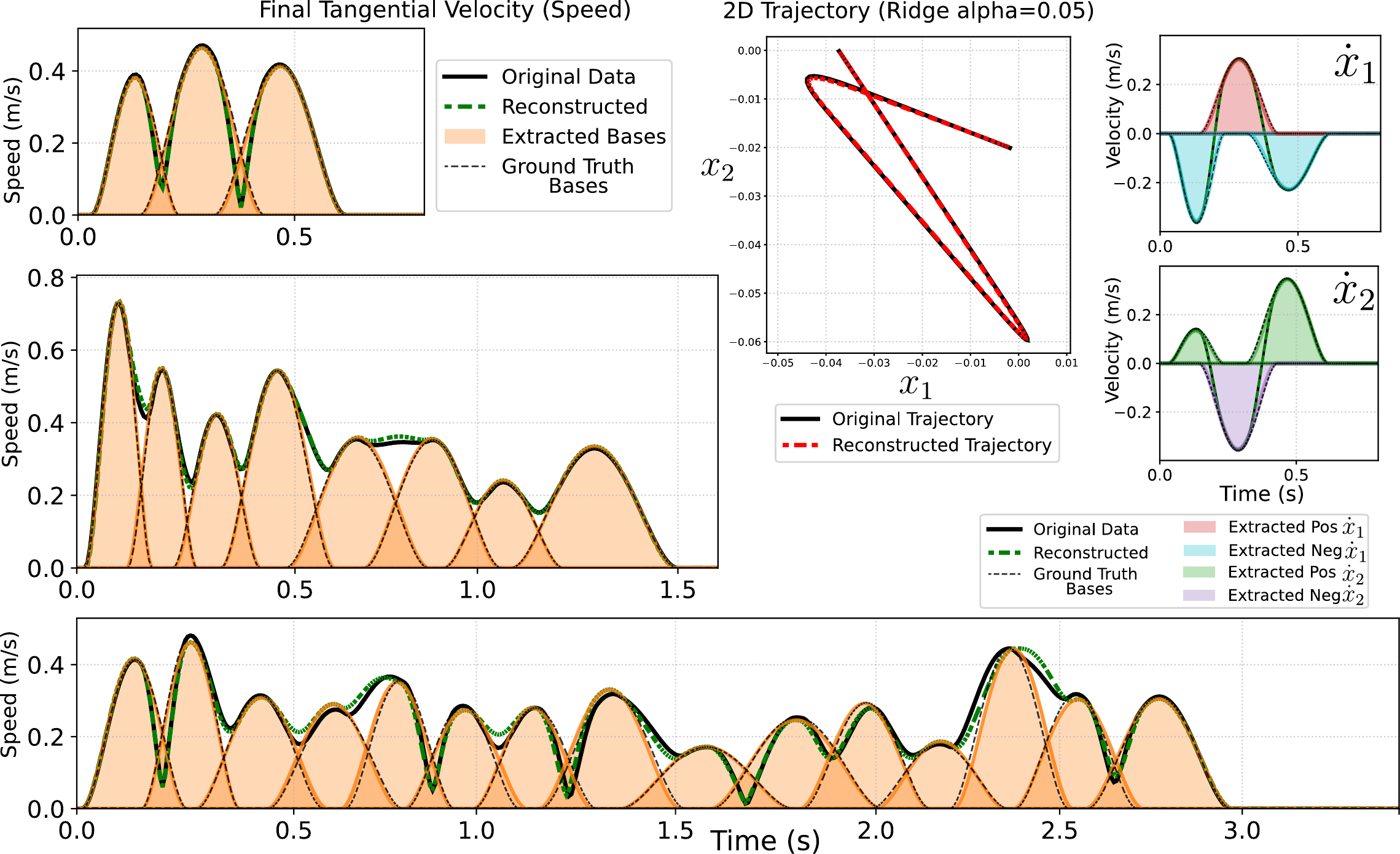}
    \caption{Example of data generated and results obtained by our method for the case of $K = 3$ (top image), $K = 8$ (middle image), and $K = 15$ (bottom image) for unimodal distributions. It can be seen that the method is capable of generating solutions that produce basis functions practically identical to those used as ground truth to generate the synthetic data. An image of the 2D trajectory for the case $K = 3$ is also included, as well as the decomposition into the $\dot x_1$ and $\dot x_2$ velocity components, where the solutions generated by our algorithm for the synthetic-data case can be observed.}
    \label{fig:Syn3815}
\end{figure}

Following this code, an example of representative decompositions at each level using synthetic data is presented (Fig.~\ref{fig:Syn3815}): the reconstructed velocity profiles closely track the ground truth (GT) data, and for $K = 3$, the recovered submovements practically coincide with those used to generate the data.

\section{Comparative Analysis of Basis Functions: Minimum Jerk vs. Lognormal Profiles}\label{MinJerkvsLGNB}
Sub-ID is agnostic to the choice of the underlying kinematic basis function. Regardless of whether Minimum Jerk (MinJerk) or Lognormal (LGNB) profiles are utilized, the extraction pipeline follows the same iterative flow. The steps explained in the main text are the same for any of the basis functions chosen for the decomposition. This makes the process highly replicable and easy to interchange depending on the basis function required by both the task and the user.

Although the structural flow remains identical, the choice between MinJerk and LGNB introduces significant differences in kinematic expressiveness (as the basis functions are defined differently for each model), optimization complexity, and numerical stability. While MinJerk defines the sub-movement as a derivative of a fourth-order polynomial parameterized by only two temporal variables, the LGNB is more complex, stemming from its ability to exhibit asymmetric velocity profiles. This asymmetry (skewness) is mapped as a Gaussian distribution over the logarithmic axis. By doing so, the basis function requires a third component to define it completely: the skewness ($\mu$). LGNBs can also require a sigma value, which represents the kurtosis ($\sigma$, representing the width) of the LGNB function. This element can also be added to the optimization process, thus introducing a fourth parameter. This difference between the parameters of the basis functions generates the most significant divergence in the execution of our algorithm. 

The inclusion of the skewness parameter ($\mu$), the kurtosis ($\sigma$), and the exponential nature of the LGNB function impact the algorithm in two critical aspects:

\noindent \textbf{A. Dimensionality of the Search Space (Grid Search Complexity):} \\
During the greedy initialization (Phase B), the algorithm searches for the optimal parameters to fit the residual. For the MinJerk model, this is a computationally inexpensive 1D search that iterates solely over candidate durations ($D$). In contrast, the standard LGNB model requires a 2D grid search, iterating over both candidate durations and a bounded space of skewness values ($\mu \in [\mu_{min}, \mu_{max}]$). Furthermore, if the kurtosis ($\sigma$) is also optimized during this phase, the search expands into a 3D grid. This exponentially increases the computational cost of the initialization phase, although it yields an initial fit of much higher fidelity.

\noindent \textbf{B. Multicollinearity and Sensitivity to Regularization:} \\
The most profound systemic difference manifests during the final spatial amplitude extraction via Ridge Regression. MinJerk bases are rigid, symmetric polynomials (i.e., their internal shape cannot be modulated independently of their temporal support). Being solely dependent on their start and end times, the optimization algorithm rarely places multiple MinJerk bases within the exact same temporal window without them being deactivated. As a result, the basis matrix $A$ is naturally well-conditioned and largely orthogonal. Consequently, the linear system does not suffer severely from multicollinearity, allowing the Ridge regularization penalty ($\alpha$) to be set to a higher value (e.g., $\alpha = 0.1$) without destroying the reconstructed shape or risking numerical instability.

\begin{figure}[ht]
    \centering
\includegraphics[width=1.0\linewidth]{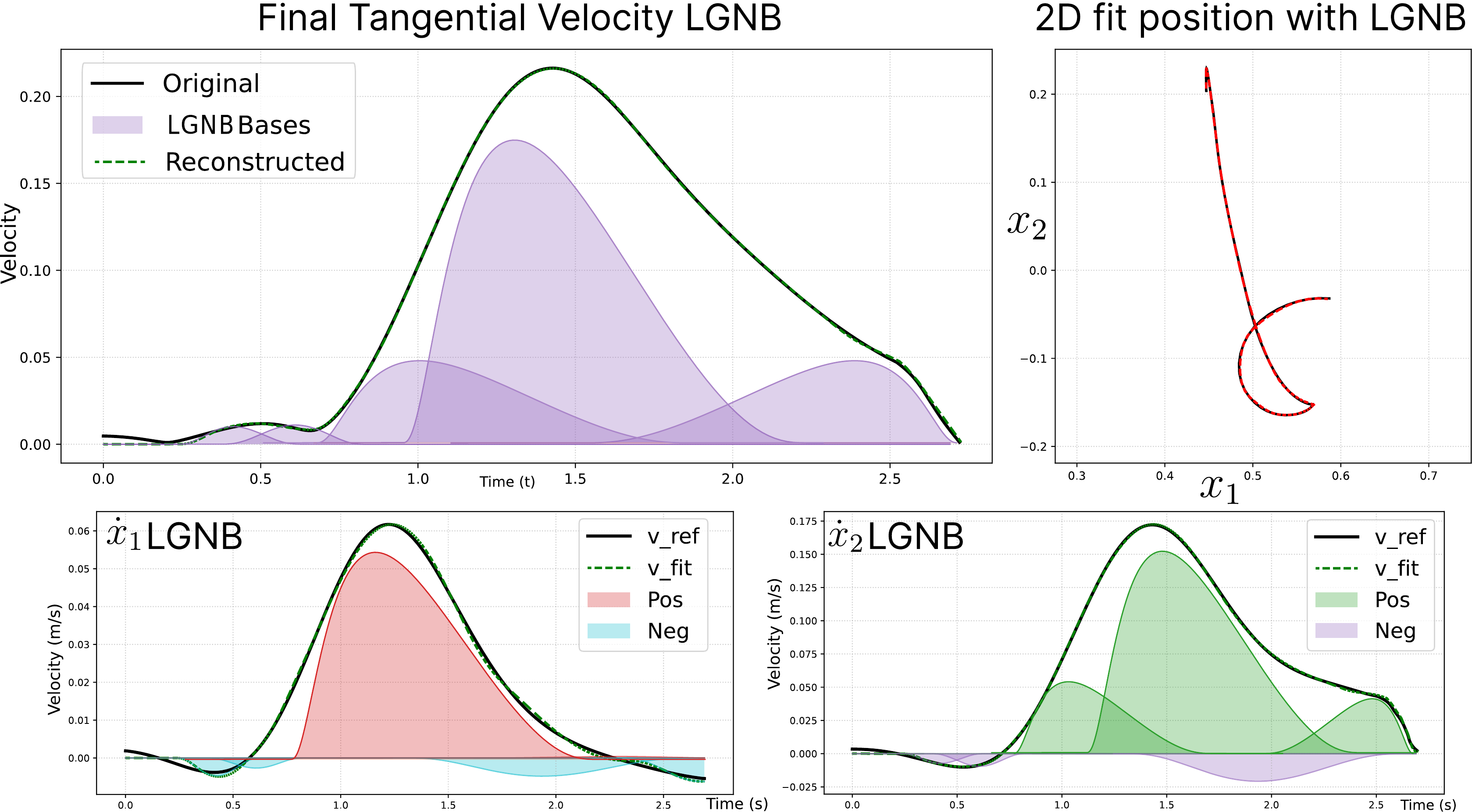}
    \caption{Trajectory fitting pipeline via LGNB with asymmetry optimization. The shaded areas represent the contribution of each optimized LGNB base. The search space for the $\mu$ parameter enables the generation of asymmetric solutions. This geometric flexibility is crucial for accurately approximating real velocity peaks skewed toward temporal boundaries, achieving a robust fit against non-symmetric dynamics and drastically reducing overfitting thanks to the coupled Ridge regularization.}
    \label{fig:NonSymetric}
\end{figure}

In contrast, while the LGNB functions are strictly bounded within $[T_0, T_1]$ just like the MinJerk bases, their parameterized flexibility allows the optimization algorithm to stack multiple bases within the exact same temporal window, differing only slightly in their skewness ($\mu$) or kurtosis ($\sigma$). These highly overlapping, intra-window bases are mathematically very similar, causing the resulting basis matrix $A$ to become highly ill-conditioned (multicollinear). Without strict regularization, the linear solver acts spuriously, assigning massive and opposing antagonist amplitudes (e.g., $+5000$ and $-4990$) to these co-located bases, which mathematically cancel each other out but lack physical meaning. However, applying an aggressive penalty would destroy the delicate linear combination of these skewed functions. This is precisely where the Adaptive-Ridge optimization resolves the sensitivity: it automatically detects this ill-conditioning and dynamically assigns a minute, highly controlled $L_2$ regularization penalty (typically finding optimal values around $\alpha \approx 5 \times 10^{-5}$). This automatically tuned penalty forces the solver to distribute realistic and bounded amplitudes among the highly correlated bases without aggressively flattening the true kinematic signal. Figure~\ref{fig:NonSymetric} presents a visual example of our method using the LGNB basis function.

\begin{figure}[ht]
    \centering
    \includegraphics[width=1.0\linewidth]{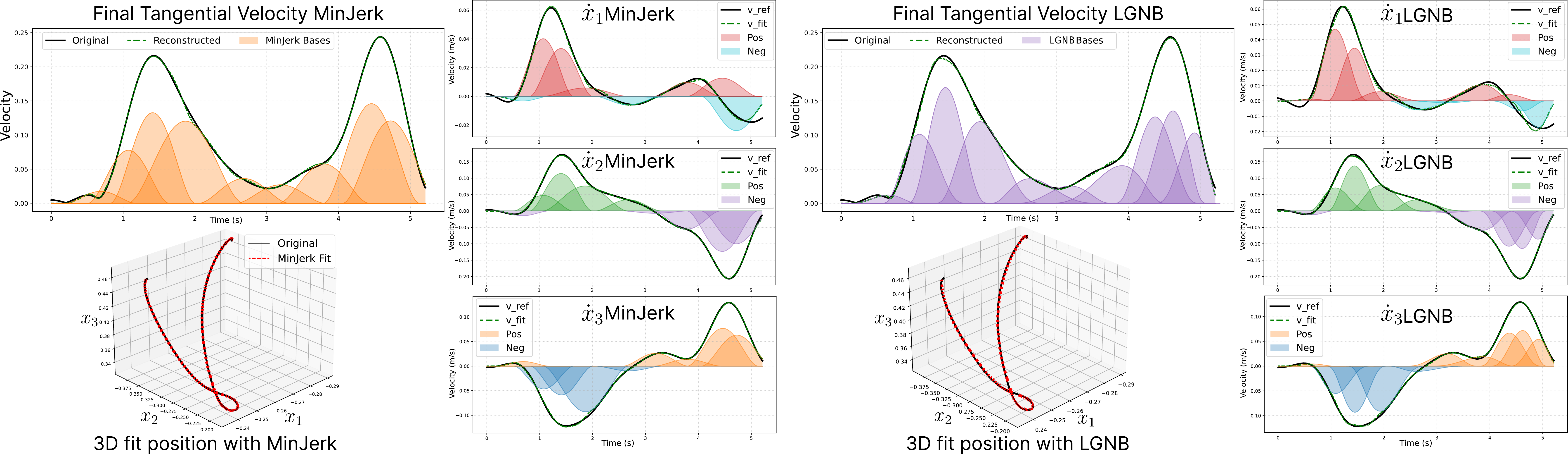}
    \caption{Comparative spatiotemporal decomposition using MinJerk and LGNB basis functions. The pipeline demonstrates its agnosticism to the underlying kinematic model by successfully decomposing the same 3D trajectory using Minimum Jerk (Left) and Bounded Lognormal (Right) profiles. Both models achieve a highly accurate 3D spatial fit (bottom left of each panel) with an approximate RMSE of $2 \times 10^{-4}$. However, the choice of basis introduces differences in optimization complexity and regularization needs. The symmetric MinJerk model (Left) resolves the trajectory using $K=9$ primitives and a Ridge penalty of $\alpha=0.1$ in $2.7$ seconds. The more complex LGNB model (Right), which optimizes additional parameters like skewness ($\mu$), requires $K=10$ overlapping primitives and a highly controlled, smaller regularization penalty ($\alpha=5 \times 10^{-5}$) to manage multicollinearity. This increased dimensionality extends the computation time to $4.0$ seconds but allows for the capture of asymmetric velocity profiles.}
    \label{fig:MinJerkvsLGNB3D}
\end{figure}
\newpage

\noindent \textbf{Empirical comparison of the algorithm's agnostic capability} \\
Figure~\ref{fig:MinJerkvsLGNB3D} presents the algorithm's agnostic capability showing the solution for the same 3D task obtained using the MinJerk method (left) and LGNB (right). In both cases, the algorithm was executed to achieve a typical error of 2\%. For this purpose, MinJerk used 9 basis functions ($K=9$), while LGNB required 10 basis functions ($K=10$).

The MinJerk solution is obtained faster ($t = 2.7\,\text{s}$), whereas LGNB generally takes longer to generate the final solution ($t = 4.0\,\text{s}$). This is due to the previously mentioned point: because of the larger grid search required when more hyperparameters are present, the algorithm needs a greater number of iterations to optimize all parameters.

For the MinJerk case, a Ridge regression with $\alpha = 0.1$ was used, while for LGNB a value of $\alpha = 5\times10^{-5}$ was employed. In both cases, the final RMSE error obtained from the position fit is approximately $2\times10^{-4}$.

\newpage


\section{Real Data Acquisition setup}\label{DataAcquisitionsetup}

\begin{figure}[ht]
    \centering
    \includegraphics[width=1.0\linewidth]{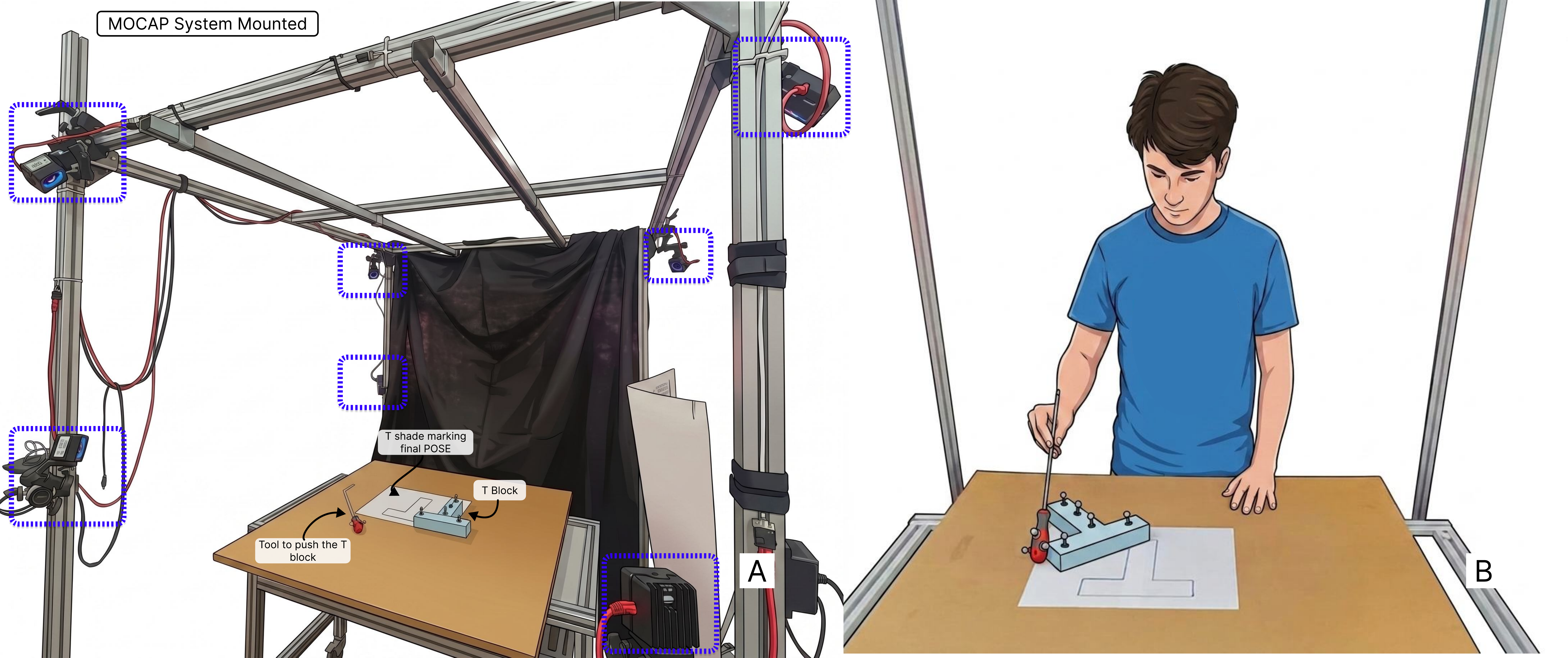}
    \caption{Experimental setup for the PushT dataset collection. (A) Overview of the complete experimental setup, including the Motion Capture (MOCAP) system used to record the tasks performed by the users. The image also shows the block to be pushed and the silhouette corresponding to its target final pose. (B) Participant performing the 2D and 3D PushT data collection tasks.}
    \label{fig:TPushCollection}
\end{figure}

During this research, the Long-Horizon 2D PushT and 3D PushT datasets were collected using an OptiTrack motion capture system equipped with eight Prime x22 cameras (Figure~\ref{fig:TPushCollection}). The system continuously recorded data through ROS1 at a sampling frequency ranging from $100\,\mathrm{Hz}$ to $120\,\mathrm{Hz}$, which was subsequently interpolated to a uniform frequency of $100\,\mathrm{Hz}$.

Simultaneously, the spatial position of the tool end-effector was recorded to directly represent the human motor output, together with the full pose (position and orientation) of the T-shaped block. To prevent loss of positional information in the presence of occlusions during the three-dimensional task, multiple markers were attached to the block.

The datasets were collected from two participants who were not involved in the development of this work. All participant information was fully anonymized, and no personal data were collected. The recorded data consisted exclusively of motion trajectories and did not contain any information that could identify the participants. Furthermore, the extracted motion data lacked any global spatial reference, making it impossible to determine the participants' location or reconstruct their movements in a global coordinate frame.

The motion capture system was entirely mounted on the tool and the T-shaped block; therefore, no sensors were attached to any of the participants. Consequently, no personal or biometric data were collected during the dataset acquisition process. 

\newpage
\section{Sub-ID quantitative results for bimodal and failure cases}


\subsection{ Bimodal distribution metrics}

Table~\ref{tab:metrics_evaluation} reports the kinematic evaluation metrics for the final solution. The synchrony error ($E_{\text{sync}}$) is $119.3\text{ ms}$ (compared to $151.4\text{ ms}$ for the Ground Truth). This temporal alignment suggests that the optimizer did not inject high-frequency noise or mathematical artifacts to compensate for the fit.

The overlapping analysis evaluates the algorithm's response to temporal aliasing in bimodal environments. Based on the temporal resolution limits ($\Delta t_{\text{min}}$), the method separates primitives by $0.0492\text{ s}$ (for $r=0.90$) and $0.0352\text{ s}$ (for $r=0.95$). Consequently, the maximum spatiotemporal correlation is $\rho=0.6595$, and the mean overlap ($0.171$) remains below the GT reference ($0.201$). Critical interactions (overlaps $>0.8$) represent $2.90\%$, a value comparable to that obtained in the unimodal case. These results quantify the algorithm's behavior in bimodal kinematic distributions, indicating that no substantial increase in structural redundancy was observed.

\begin{table}[htbp]
    \centering
    \caption{Kinematic and Mathematical Identifiability Metrics for Bimodal Synthetic Data}
    \label{tab:metrics_evaluation}
    \begin{tabular}{lcc}
        \toprule
        \textbf{Metric} & \textbf{Reference} & \textbf{Sub-ID} \\
        \midrule
        Sync Error ($E_{\text{sync}}$) & 151.4 ms & 119.3 ms \\
        $\Delta t_{\text{min}}$ ($T_{\text{med}}$, $r=0.95$) & --- & 0.0352 s \\
        $\Delta t_{\text{min}}$ ($T_{\text{med}}$, $r=0.90$) & --- & 0.0492 s \\
        Max Spatiotemporal $\rho$ & --- & 0.6595 \\
        Mean Overlap ($\mu \pm \sigma$) & 0.201 $\pm$ 0.021 & 0.171 $\pm$ 0.036 \\
        Overlaps $> 0.8$ ($\mu \pm \sigma$) & 0.00\% $\pm$ 0.00\% & 2.90\% $\pm$ 1.64\% \\
        \bottomrule
    \end{tabular}
\end{table}

\subsection{ Failure case metrics}
Table~\ref{tab:metrics_evaluationFail} quantifies the failure of the Sub-ID method and evidences the overall degradation of the reconstruction compared to the GT, indicating the introduction of severe artifacts and kinematic distortions to force the fit.

The overlapping metrics confirm the collapse of the compositional structure. The mean overlap doubles compared to the reference (increasing from $0.251$ to $0.534$), a pattern repeated in the percentage of critical interactions (overlaps $>0.8$), which rises to $33.90\%$. This failure analysis empirically demonstrates the operational boundary of our optimization framework: if the theoretical limits of spatiotemporal resolution are exceeded, the inverse problem becomes inherently ill-conditioned, making the exact recovery of the latent parameters impossible.
\begin{table}[htbp]
    \centering
    \caption{Kinematic and Mathematical Identifiability Metrics for Unimodal Failure Synthetic Data}
    \label{tab:metrics_evaluationFail}
    \begin{tabular}{lcc}
        \toprule
        \textbf{Metric} & \textbf{Reference} & \textbf{Sub-ID} \\
        \midrule
        Sync Error ($E_{\text{sync}}$) & 297.4 ms & 212.3 ms \\
        $\Delta t_{\text{min}}$ ($T_{\text{med}}$, $r=0.95$) & --- & 0.0449 s \\
        $\Delta t_{\text{min}}$ ($T_{\text{med}}$, $r=0.90$) & --- & 0.0629 s \\
        Max Spatiotemporal $\rho$ & --- & 0.4982 \\
        Mean Overlap ($\mu \pm \sigma$) & 0.251 $\pm$ 0.031 & 0.534 $\pm$ 0.029 \\
        Overlaps $> 0.8$ ($\mu \pm \sigma$) & 15.00\% $\pm$ 5.76\% & 33.90\% $\pm$ 8.08\% \\
        \bottomrule
    \end{tabular}
\end{table}

\section{Mathematical Details: Identifiability Analysis}
Table~\ref{tab:evaluation_metrics_short} presents a summary of all the metrics, along with the associated papers where these metrics are described in a mathematically detailed manner. A more detailed mathematical description for some of the metrics are also presented in this Section.
\begin{table}[htbp]
\centering
\small
\caption{\textbf{Summary of Statistical Analysis and Evaluation Metrics.} Mathematical formulations used to assess kinematic fidelity and structural identifiability.}
\label{tab:evaluation_metrics_short}
\renewcommand{\arraystretch}{1.8}
\begin{tabular}{l c c}
\toprule
\textbf{Metric} & \textbf{Equation} & \textbf{Reference} \\
\midrule

\textbf{Pos Error (RMSE)} & 
$\text{RMSE}_{\text{pos}} = \sqrt{\frac{1}{N} \sum_{n=1}^{N} \left\| P(t_n) - \hat{P}(t_n) \right\|^2}$ & 
- \\

\textbf{Vel $R^2$} & 
$R^2_{\text{vel}} = 1 - \frac{\sum (v(t_n) - \hat{v}(t_n))^2}{\sum (v(t_n) - \bar{v})^2}$ & 
- \\

\textbf{Vel RMSE} & 
$\text{RMSE}_{\text{vel}} = \sqrt{\frac{1}{N} \sum_{n=1}^{N} \left( v(t_n) - \hat{v}(t_n) \right)^2}$ & 
- \\

\textbf{Primitive Rate} & 
$\text{Rate} = \frac{K}{T_{\text{total}}}$ & 
- \\

\textbf{Temporal Overlap} & 
$O_{i,j} = \frac{\text{intersect}(t_{i}, t_{j})}{\min(T_i, T_j)}$ & 
\cite{rohrer2003} \\

\textbf{Spatiotemporal Kernel Correl.} & 
$\rho_{i,j} = \cos(\psi_{i,j}) \cdot \rho_{i,j}^{\text{temp}}$ & 
\cite{rohrer2003, donoho2003optimally} \\

\bottomrule
\end{tabular}
\end{table}

\subsection{Kernel Correlation and Identifiability Limit}

As established in Equation 17 of the main manuscript, the spatiotemporal kernel correlation between two distinct submovements $i$ and $j$ is given by:
\begin{equation}
  \rho_{i,j} = \cos(\psi_{i,j}) \cdot \rho_{i,j}^{\text{temp}}
\end{equation}

To analyze the identifiability limit due to temporal overlap, let us consider two collinear submovements ($\cos(\psi_{i,j}) = 1$) that share the same duration $T$ and shape parameters, differing only by a temporal shift $\Delta t$ (i.e., $t_{0,j} = t_{0,i} + \Delta t$). Their temporal correlation simplifies to a function of the time-shift and duration:
\begin{equation}
  \rho(\Delta t, T) = \frac{\langle \phi_{\text{vel}}(t; \theta_i), \phi_{\text{vel}}(t; \theta_j) \rangle}{\|\phi_{\text{vel}}(\cdot; \theta_i)\|_2 \|\phi_{\text{vel}}(\cdot; \theta_j)\|_2}
\end{equation}

Properties: $\rho(0, T) = 1$; $|\rho(\Delta t, T)| \le 1$. When $\rho(\Delta t, T) \approx 1$, two kernels separated by $\Delta t$ are nearly collinear, making the decomposition ill-conditioned.

A practical identifiability criterion chooses a threshold $\rho_{\text{max}}$ (e.g., 0.9 or 0.95) and defines the minimum resolvable separation:
\begin{equation}
  \Delta t_{\text{min}}(T) \triangleq \inf\{\Delta t > 0 : \rho(\Delta t, T) \le \rho_{\text{max}}\}
\end{equation}

\subsection{Scale Invariance for Minimum-Jerk Kernels}

Instead of redefining the Minimum-Jerk velocity pulse, we rely on its definition from the main text, where it is fully parameterized by its duration $T$. By substituting the normalized time shift $\xi = \Delta t / T$ and the intermediate normalized time variable $u = (t - t_{0,i}) / T$, the temporal correlation integral becomes:
\begin{equation}
  \rho(\xi) = \frac{\int_\xi^1 s(u) s(u-\xi) du}{\int_0^1 s(u)^2 du}
\end{equation}
where $s(u)$ represents the standard shape polynomial of the Minimum-Jerk profile. 

The correlation depends only on the ratio $\xi$, not on $T$ separately. Consequently, $\Delta t_{\text{min}}(T) = T \cdot \xi_{\text{min}}$ scales linearly with $T$: longer primitives require a larger onset separation to remain identifiable.

\subsection{Spatiotemporal Generalization}

In $\mathbb{R}^D$, substituting the true spatiotemporal inner product back into our normalized formulation yields the complete kernel correlation (as shown in the main text):
\begin{equation}
  \rho_{i,j}(\Delta t, T) = \cos(\psi_{i,j}) \cdot \rho(\Delta t, T)
\end{equation}

where $\psi_{i,j}$ is the spatial angle between primitive directions. When the spatial vectors are orthogonal ($\cos(\psi_{i,j}) = 0$), $\rho_{i,j} = 0$, and the primitives are perfectly identifiable regardless of their temporal overlap. This mathematically reveals that 3D motor tasks extend identifiability well beyond the 1D temporal bound.

\subsection{Algorithmic Implementation in Multi-Dimensional Spaces}

To operationalize the spatiotemporal generalization in $D$-dimensional tasks (e.g., $D=2$ or $D=3$), the Sub-ID algorithm decouples the estimation of the shared temporal parameters from the spatial directions. The procedure is implemented in three main steps: First, the temporal parameters (e.g., onset times and durations) are initialized by analyzing the scalar speed profile $v_t(t) = \|\mathbf{v}(t)\|_2$. This dimension-agnostic approach allows the algorithm to robustly detect submovements regardless of their spatial orientation. Second, a shared temporal basis matrix $\Phi \in \mathbb{R}^{N \times K}$ is constructed for all $K$ detected submovements. The temporal support of each submovement is strictly identical across all spatial dimensions. Finally, rather than optimizing spatial angles directly, the algorithm recovers the spatial directions and amplitudes by solving $D$ independent Ridge regressions. For each spatial coordinate axis $d \in \{1, \dots, D\}$, a weight vector $\mathbf{w}_d \in \mathbb{R}^K$ is estimated. For any given submovement $k$, its total spatial amplitude $w_k$ and unit direction vector $\mathbf{e}_k$ are subsequently recovered as:
\begin{equation}
  w_k = \sqrt{\sum_{d=1}^D w_{d,k}^2}, \qquad \mathbf{e}_k = \frac{1}{w_k} [w_{1,k}, \dots, w_{D,k}]^\top.
\end{equation}
This approach ensures that the multidimensional decomposition remains computationally efficient while implicitly capturing the spatial angles $\psi_{i,j}$ that govern spatiotemporal identifiability.

\bibliography{mainOverleaf}